\documentclass{article}

\usepackage{PRIMEarxiv}

\usepackage{lipsum}
\usepackage{hyperref}
\usepackage{url}

\usepackage{graphicx}
\usepackage{wrapfig}                                                                                                                           \usepackage{cancel}
\usepackage{subfigure}
\usepackage{enumerate}
\usepackage{enumitem}
\usepackage{amssymb,amsmath}
\usepackage{epstopdf}
\usepackage{epsfig}
\usepackage{textcomp}                                                                                                                          \usepackage{sidecap}
\usepackage{floatrow}
\usepackage{enumitem}
\usepackage{changepage}   
\usepackage{enumitem}
\usepackage{mathrsfs}
\usepackage{amsfonts}
\usepackage{xfrac}
\usepackage{algpseudocode,algcompatible}
\usepackage[linesnumbered,ruled,vlined]{algorithm2e}
\DontPrintSemicolon
\renewcommand{\COMMENT}[2][.5\linewidth]{%
  \leavevmode\hfill\makebox[#1][l]{//~#2}}
\algnewcommand\algorithmicto{\textbf{to}}
\algnewcommand\RETURN{\State \textbf{return} }

\usepackage{floatrow}
\newfloatcommand{capbtabbox}{table}[][\FBwidth]
\usepackage{makecell}
\title{Fuse-PINN: Combining Finite Element Methods and Physics-Informed Neural Networks to enable massively parallel simulations of non-linear PDEs}

\author{S.J.P.Pamela\thanks{corresponding author: stanislas.pamela@ukaea.uk}\\
UKAEA, CCFE, Culham Science Centre,\\Abingdon, Oxon, OX14 3DB, UK.
\And
N.Pailliez\\
IRMA, Univ. Strasbourg, France.\\
\And
E.Franck\\
IRMA, Univ. Strasbourg, France.\\
INRIA Nancy-Grand Est, Strasbourg, France.
\And
V.Michel-Dansac\\
IRMA, Univ. Strasbourg, France.\\
INRIA Nancy-Grand Est, Strasbourg, France.
\And
L.Navoret\\
IRMA, Univ. Strasbourg, France.\\
INRIA Nancy-Grand Est, Strasbourg, France.
\And
V.Gopakumar\\
UKAEA, CCFE, Culham Science Centre, \\Abingdon, Oxon, OX14 3DB, UK.\\
}

\DeclareMathAlphabet{\mathbfcal}{OMS}{cmsy}{b}{n} 

\newcommand{\be}{\begin{equation}}
\newcommand{\ee}{\end{equation}}
\newcommand{\ba}{\begin{eqnarray}}
\newcommand{\ea}{\end{eqnarray}}
\newcommand{\ttt}{\texttt}

\begin{document}
\maketitle

\begin{abstract}
Research in the domain of Physics-Informed Neural Networks applied to PDE modelling has advanced significantly in recent years, providing a wide ecosystem of novel methods. However, one frequent limitation is the inability of PINNs to scale out on High-Performance Computing architectures to address more ambitious challenges, particularly non-linear physics models with turbulent dynamics in large spatial domains. Typical studies restrain demonstrations to toy models with simple physics equations, few variables, and rarely any complex non-linear dynamics evolved over long timescales. Part of the AI community has even adopted the idea that ``PINNs don't scale''. This work proves otherwise. We present a mathematically robust formulation of Finite-Element-Method for Neural Networks (FEM-NN) implemented into a comprehensive, modular framework: Fuse-PINN. This formulation provides the basis for a scalable parallelisation of Physics-Informed Neural Networks on GPU-based HPC systems. Furthermore, an Energy Natural Gradient Descent optimiser is implemented to allow high-precision convergence. Fuse-PINN is demonstrated on physics models relevant to Magnetic Confinement Fusion, including non-linear convective models with fine-scale dynamics. Near-perfect scaling is achieved on multi-GPU runs with increasing problem size.
\end{abstract}



\section{Introduction}

\subsection{Motivation}

The numerical modelling of non-linear systems of partial differential equations (PDEs) is used in research and industrial applications across scientific and engineering challenges. From aerospace to automotive, from weather to climate predictions, from solar to fusion energy, numerical solvers are a core component of engineering design and technological advances. Predicting the non-linear evolution of complex physical systems typically relies on conventional numerical methods, which are constantly being developed and improved, both for academic scientific research (\cite{MOOSE,MFEMweb,Firedrake,OpenFoam,JOREK}) as well as industrial tools (\cite{ANSYS,ABAQUS,SIEMENS}). In recent years, a particular branch of Machine Learning has brought a promising alternative to these conventional approaches: Physics-Informed Neural Networks \cite{Raissi2019}. PINNs train a Neural Network to learn the solution of a PDE by using the residual of the equations directly, without requiring training data from other simulations. Following its initial demonstration, this field of study has now evolved into a vast array of numerical recipes, providing a range of building blocks that may, in theory, be assembled to create powerful simulation frameworks. However, to the best of our knowledge, a practical implementation of PINN methods has yet to be demonstrated on modern, multi-node, multi-GPU High-Performance-Computing (HPC) architectures, with ambitious applications requiring complex non-linear physics systems. This work is inspired by the pioneering achievements of PINN researchers, to combine these various building blocks and deliver a scalable formulation of Finite Element Methods for Neural Networks (FEM-NN).

Three practical barriers continue to limit the use of PINNs for challenging scientific simulations. First, high-accuracy training remains difficult because the loss landscape is often poorly conditioned, with competing domain, initial and boundary residual terms that can converge different rates. Second, many physically relevant non-linear PDE systems contain multiple length and time scales, with sharp gradients, transport-dominated dynamics, and long time evolution, all of which expose the spectral bias and optimisation fragility of standard neural-network approximations. Third, although conventional simulation codes routinely exploit distributed-memory HPC systems, many PINN methods remain confined to single-GPU implementations, limiting the size of the domain, the number of collocation points, and the complexity of the physics that can be treated in practice.

We propose Fuse-PINN \cite{Fuse-PINN-github}: a scalable framework for Physics Informed Neural Networks that combines finite-element residual assembly with a physics-based high-precision optimiser, demonstrated on High-Performance Computing (HPC) architectures. The implementation is designed for multi-node, multi-GPU execution, with a domain decomposition embedded directly into the FEM-NN formulation. We demonstrate the approach on nonlinear models relevant to fusion energy, including convective systems with fine-scale dynamics, as a step towards PINN methods that can operate beyond the limits of single-device PINN workflows.
\begin{figure}[ht!]
  \centering
        \subfigure[ ]{\includegraphics[width=0.315\linewidth]{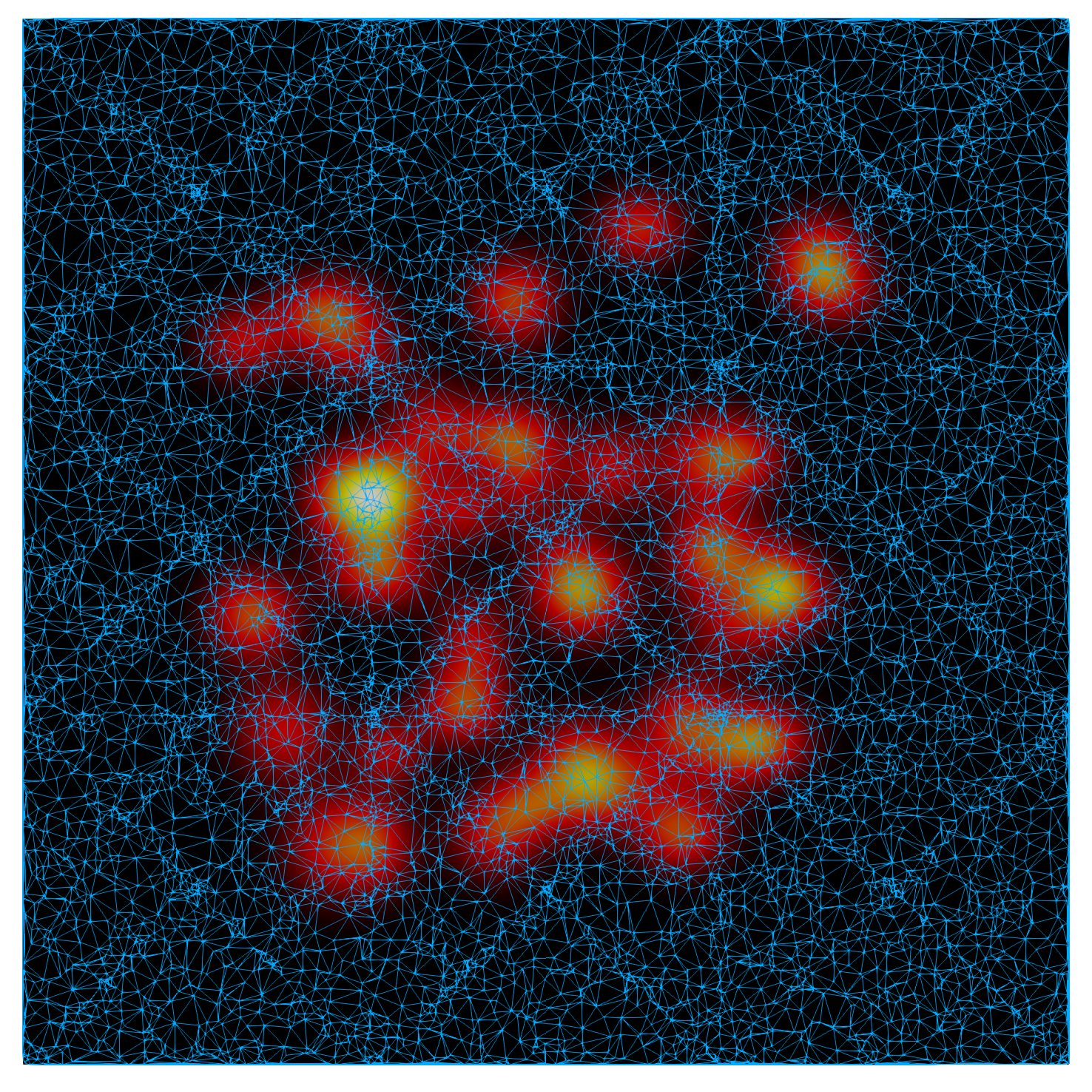}}
        \subfigure[ ]{\includegraphics[width=0.660\linewidth]{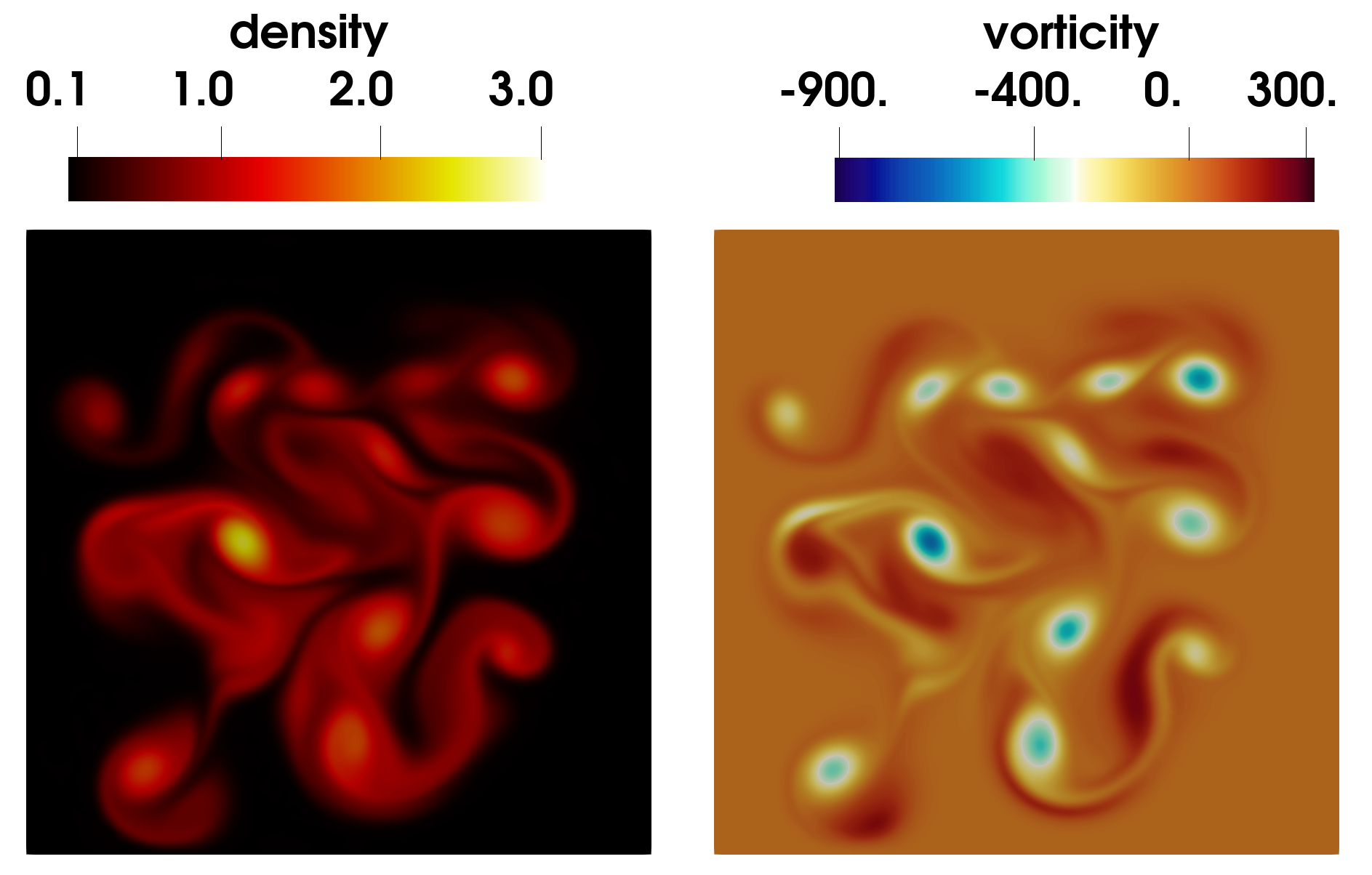}}
  \caption{\textnormal{(a)} Example of a Fuse-PINN grid which can be built directly with \cite{GMSHweb}. Intentionally low resolution sampling (blue grid) is used to illustrate the parallelisation method. The slightly higher sampling near FEM-NN edges highlights the 25 FEM-NN nodes and 36 elements. \textnormal{(b)} Simulation of Hasegawa-Wakatani demonstrated using 75 NN's parallelised on 75 GPUs (3 physics variables for 25 FEM-NN nodes).}
  \label{Figure_intro}
\end{figure}

\subsection{Background and Related Work}

\textbf{Complex Physics Models:} PINNs have been applied to a wide range of non-linear PDEs, including Burgers-type systems, diffusion equations, Navier-Stokes flows and wave propagation \cite{Raissi2019,Wang2025}. These studies have demonstrated the conceptual flexibility of physics-informed learning: the same framework can incorporate governing equations, initial and boundary conditions, sparse measurements, unknown parameters, and differentiable surrogate quantities. However, most PINN demonstrations remain concentrated on relatively low-dimensional benchmarks, simplified geometries, or short-time dynamics, with some notable exceptions, such as 3D turbulence simulations \cite{Wang2025} (and references therein, like PirateNets \cite{Wang2024}). The physics models relevant to fusion research can be most challenging: transport, instability, and turbulence models involve stiff non-linear couplings, scale separation, anisotropic dynamics, high-order derivatives, while requiring long integration windows. For instance, the MHD equations involve 8 variables (3D velocity and magnetic fields, plus density and temperature), while gyrokinetic turbulence evolve a density distribution in 5D.

\textbf{High-Precision Optimisers:} Among several novel methods in high-precision PINN optimisation is the recent work of \cite{Muller2023}. Standard optimisers like Adam and L-BFGS can provide useful approximations, but they typically fail to reach the level of precision expected from mature numerical solvers, particularly for stiff or non-linear PDEs due to the ill-conditioning of PINNs \cite{Rathore2024,Cao2024}. M\"uller \& Zeinhofer's Energy Natural Gradient Descent method (ENGD) addresses this limitation by defining the optimisation step in function space using a physics-based preconditioner, so that the update direction approximates a Newton step projected onto the tangent space of the neural-network manifold. This method achieves errors orders of magnitude lower than standard optimisers on representative benchmark problems. Notable extensions of ENGD as well as other high-precision approaches include \cite{Guzman-Cordero2025,Webb2026,Schwencke2025}.

\textbf{Domain Decomposition:} Domain decomposition has emerged as one of the most natural routes for improving both the trainability and scalability of PINNs. Conservative PINNs (cPINNs) and extended PINNs (XPINNs) decompose the physical domain into subdomains, assigning separate neural networks to each region and enforcing interface consistency through additional constraints \cite{Jagtap-cPINNs,Jagtap-XPINNs}. XPINNs generalise this idea to flexible decompositions in space and time, enabling different architectures, sampling densities, and optimisation settings in different subdomains \cite{Jagtap-XPINNs}. Finite Basis PINNs (FBPINNs) provide a related construction inspired by finite-element methods, in which compactly supported neural components are combined over overlapping subdomains, reducing spectral-bias effects through local input normalisation and smaller subdomain networks \cite{Moseley2021}. These approaches show that locality is beneficial, but they also reveal an open question: how to combine domain decomposition with a high-precision optimiser and a finite-element residual formulation in a way that maps cleanly onto modern distributed GPU systems. The works by \cite{Mitusch2021} and \cite{Meethal2022} also propose FEM-NN ideas, but these are conceptually different from the nodal NN formulation we present here. Likewise, hp-VPINN \cite{Kharazmi2020} combines Galerkin discretisation with a single PINN network, but it does not provide a recipe for parallelisation. Finally, TensorGalerkin \cite{Wen2026} augments a standard Galerkin/FEM discretisation with differentiable assembly and neural-operator learning, but it does not provide a parallelisation recipe for pure PINNs.

\textbf{Multi-node, Multi-GPU PINNs:} Compared with the extensive literature on PINN formulations, relatively few studies have demonstrated PINNs as genuinely distributed high-performance computing workloads. Existing work on parallel cPINNs and XPINNs has shown that MPI-based domain decomposition can enable CPU and GPU parallelism, with strong and weak scaling studies on selected forward and inverse problems \cite{Shukla2021}. To the best of our knowledge, \cite{Shukla2021} and \cite{Ohashi2026} are the only demonstration of PINN parallelisation onto multi-GPUs using domain decomposition. FBPINN methods could, in theory, enable parallelisation, but that has not yet been clearly demonstrated at scale \cite{Moseley2021,Dolean2024}. These results are encouraging, but they do not fully resolve the requirements of large-scale scientific modelling: the method must support local physics assembly, efficient communication, stable high-precision optimisation, and nonlinear multi-field residuals. In particular, there remains a gap between proof-of-principle distributed PINNs and a framework that can treat fusion-relevant PDE systems at scales where single-GPU memory and runtime constraints become prohibitive.

\textbf{Fuse-PINN:} We propose a PINN framework that combines FEM-NN and ENGD to demonstrate scalability on HPC architectures. The implementation is designed for multi-node, multi-GPU execution, with a domain decomposition embedded directly into the FEM-NN formulation. The finite-element structure provides locality, minimising communications while retaining the ability to resolve global solutions, essential for implicit time-schemes. ENGD provides a physics-aware optimisation strategy designed to reach higher-precision solutions than standard first-order methods. The domain decomposition is distributed across devices while maintaining the global physics constraints required by the PDE system. We demonstrate the approach on fusion-relevant nonlinear models, including convective systems with fine-scale dynamics, as a step towards physics-informed simulation methods that can operate beyond the limits of single-device PINN workflows.

\section{Methods}

\begin{figure}[ht!]
  \centering
        \includegraphics[width=0.98\linewidth]{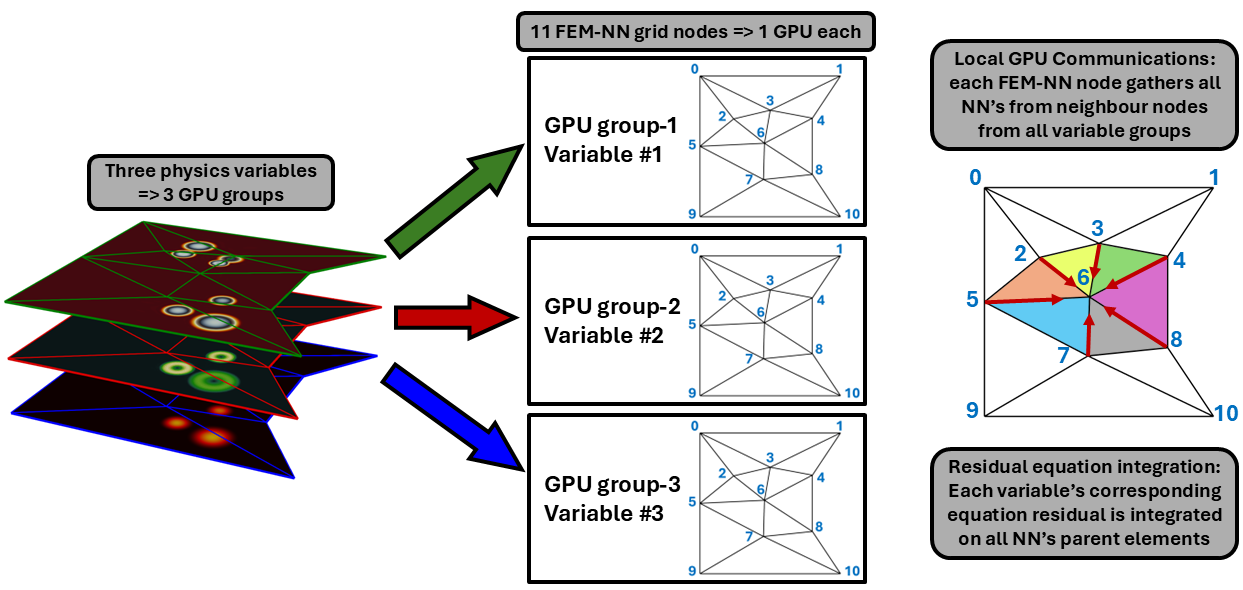}
  \caption{Parallelisation strategy of the Fuse-PINN implementation with FEM-NNs. Each physics variable (and its corresponding equation) is allocated to a GPU group, and within that group each FEM-NN grid node is allocated to a GPU, such that each Neural Network (on a given GPU) represents the solution of one variable on all parent elements of that FEM-NN grid node.}
  \label{FIGURE_parallelisation}
\end{figure}

A fully detailed description of the PyTorch implementation of Fuse-PINN and its numerical methods is included in Appendix-\ref{AppendixA}, with particular attention to non-trivial details that are required for a successful implementation and numerical stability. Some of these details are effectively novel and undocumented elsewhere since this approach is, to the best of our knowledge, the first of its kind. The key components of the Fuse-PINN framework are:

\begin{enumerate}[itemsep=0mm]
    \item \textbf{Discrete PINNs.} The NN optimisation is done one time-step at a time. The initial conditions are treated as a single step for $t=0$, and subsequently each solution of a given time-step is used as the initial guess for the next step. Thus, if time-steps are small, the initial guess should already be close to the target solution.
    \item \textbf{Implicit time discretisation.} Time-stepping is done fully implicitly with the Crank-Nicolson scheme. It ensures numerical stability and allows large time-steps, independent of the spatial resolution. The time-step size is typically governed by the PDE dynamics.
    \item \textbf{Finite-Element Method for Neural Networks (FEM-NN).} The spatial domain is decomposed using generic triangular elements with nodal NN's, akin to finite-element-methods. This provides a local description of the solution on each element, reducing the amount of communications needed, while retaining the FEM-NN's ability to communicate dynamics across the entire grid, which is essential for implicit time-schemes. A purpose-built barycentric weight function (or basis function) is designed to ensure continuity of the solution and its derivatives across element edges.
    \item \textbf{Segregated Neural Networks.} Separate NN's are used for each physics variable. This was found to be numerically more robust, as it avoids competition between different equation losses inside a single NN loss. This approach is also aligned with our aim to enable parallel scalability, as it allows parallelisation across variables as well as FEM-NN nodes.
    \item \textbf{Energy Natural Gradient Descent optimiser (ENGD).} The physics-based optimiser is essential for faster solutions as well as higher precision. The modular implementation is such that users only have to provide the physics equations without having to worry about the ENGD implementation itself, allowing for rapid development of new physics models.
    \item \textbf{Multi-node, multi-GPU parallelisation.} The parallelisation is embedded into the FEM-NN domain decomposition, providing scalability towards large resolutions. Tests were run on Nvidia-H100's and AMD-Mi300X's. Figure-\ref{FIGURE_parallelisation} illustrates Fuse-PINN's parallelisation architecture, with two levels of distribution: across variables and across FEM-NN nodes.
\end{enumerate}

\section{Experiments}

\subsection{Basic Tests for ENGD and FEM-NN Implementations}\label{Section_vorticity_twin_blobs}

The first basic validation tests focus on the ENGD implementation, employing the Poisson model with a single variable $\Phi$ -- model (iv) in Appendix-\ref{AppendixB_Basic} -- to solve for the same field as in \cite{Muller2023}, namely $\nabla^2\Phi = -2\pi^2 sin(\pi x)sin(\pi y)$, with zero boundary conditions on a square domain with $x\in[0,1]$ and $y\in[0,1]$. This has an exact analytical solution of $\Phi = sin(\pi x)sin(\pi y)$. The test is conducted with a single PINN for the whole domain (ie. without FEM-NN) and compared against M\"uller's \& Zeinhofer's JAX implementation in \cite{Muller-github}. Details of these tests are given in Appendix-\ref{AppendixC}, and they are broadly satisfactory.

\begin{figure}[ht!]
  \centering
        \subfigure[ ]{\includegraphics[height=0.35\linewidth]{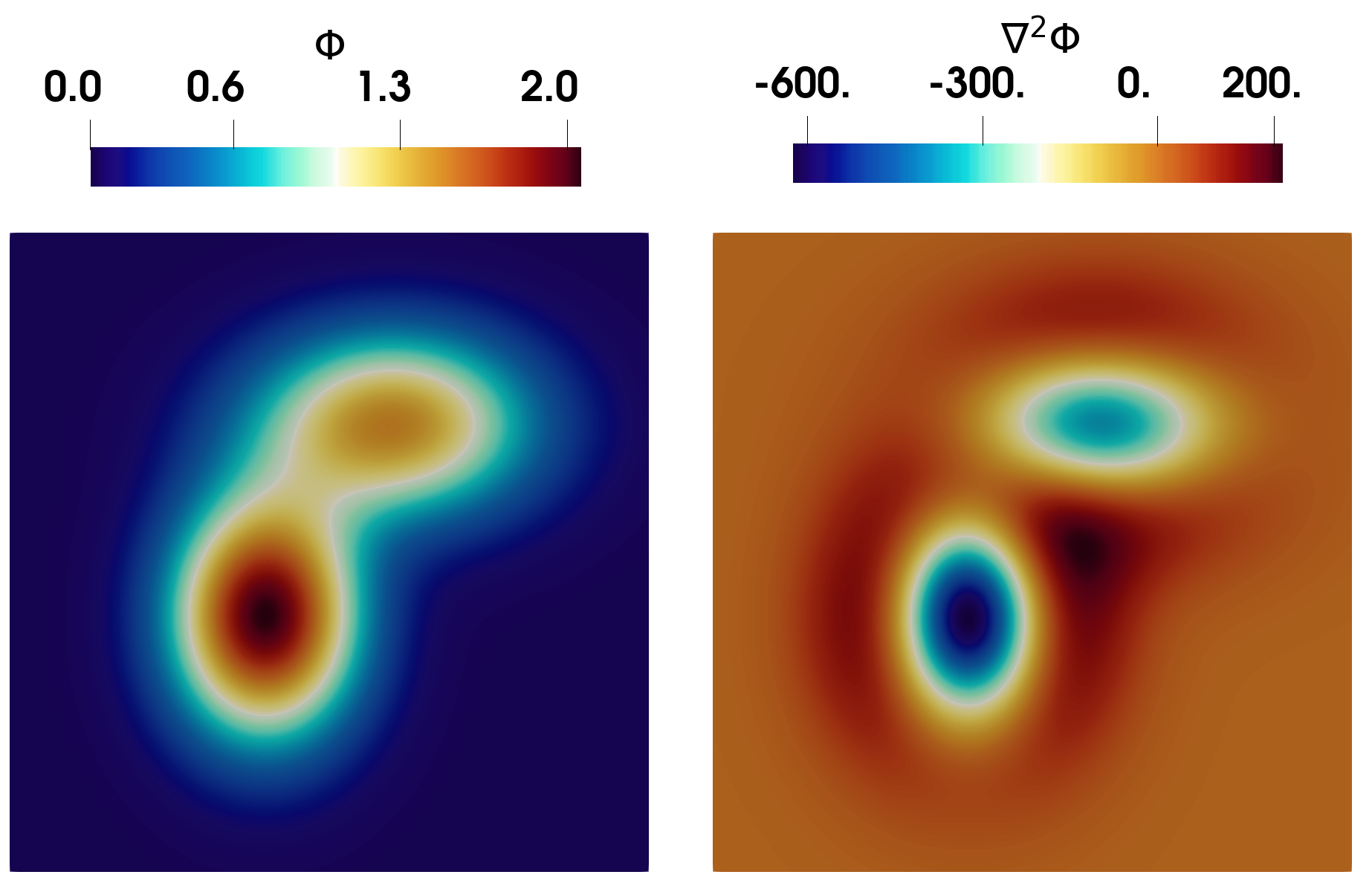}}
        \subfigure[ ]{\includegraphics[height=0.35\linewidth]{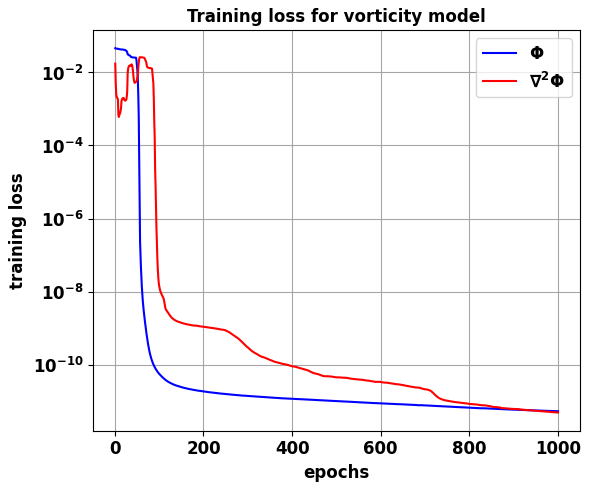}}
  \caption{Testing of the vorticity model with 2 variables. \textnormal{(a)} The 2D maps of the fields $\Phi$ and $\nabla^2\Phi$. \textnormal{(b)} The training loss as a function of epochs for each variable.}
  \label{FIGURE_Vorticity}
\end{figure}

The vorticity model with 2 variables -- model (iii) in Appendix-\ref{AppendixB_Basic} -- is a basic model used for testing code developments, particularly with parallelisation of multiple variables and multiple FEM-NN grid nodes. As shown in Figure-\ref{FIGURE_Vorticity}a, the test solves for $\Phi$ to learn a simple distribution with two asymmetric Gaussians, and for the corresponding $\nabla^2\Phi$ field. Unlike the Poisson model, for this particular case, boundary conditions are not needed since the model is entirely explicit.

This test is conducted here on a square domain with $x\in[10,11]$ and $y\in[-0.5,0.5]$. The grid is constructed with 9 nodes (8 elements), using a total sampling of 200$\times$200. Each FEM-NN node has [2,10,10,1] layers, so 2$\times$9$\times$151=2,718 parameters in total. Figure-\ref{FIGURE_Vorticity}b shows the training loss evolution for both models $\Phi$ and $\nabla^2\Phi$.

Note that, for such a simple $\Phi_0$ field, it is not at all necessary to use 9 FEM-NN nodes, and in fact using a single PINN network is more efficient. Multiple FEM-NN nodes are used here for demonstrative purposes. More quantitative evaluations of the FEM-NN scaling efficiency are presented in Section-\ref{Section_scalings}.

\subsection{Hasegawa-Wakatani model}

The first advanced, time-evolutive, non-linear model implemented in Fuse-PINN is the so-called Hasegawa-Wakatani model \cite{Hasegawa1983}. This is a reduced, electrostatic, fluid turbulence model in the presence of a strong magnetic field, thus relevant to magnetic confinement fusion. Despite its simplicity, it remains a foundational model for understanding turbulence in tokamak plasmas, and has been studied extensively since its publication \cite{Leddy2016,Gurcan2022,Artigues2025,Castagna2024}.

The Hasegawa-Wakatani model solves for 2 physics variables, the density $\rho$ and the electric potential $\Phi$, with a third auxiliary variable $w$ to reduce the order of direct derivatives needed in the system. The model is described in details in Appendix-\ref{AppendixB_HW}.

The first Hasegawa-Wakatani simulation is run in a 2D square domain with the similar initial conditions as in Figure-\ref{FIGURE_Vorticity}. The grid, with $x\in[10,11]$ and $y\in[-0.5,0.5]$, uses a resolution of 13 FEM-NN nodes, with a total sampling of 640$\times$640 to ensure high resolution of the thin vorticity layers resulting from the convection cells. The simulation is evolved for 240 time-step of $5.10^{-4}s$. Each FEM-NN node evolves a NN of 1981 parameters for each variable. The perpendicular diffusion and viscosity are set to $10^{-4}$.

\begin{figure}[ht!]
  \centering
        \includegraphics[width=0.98\linewidth]{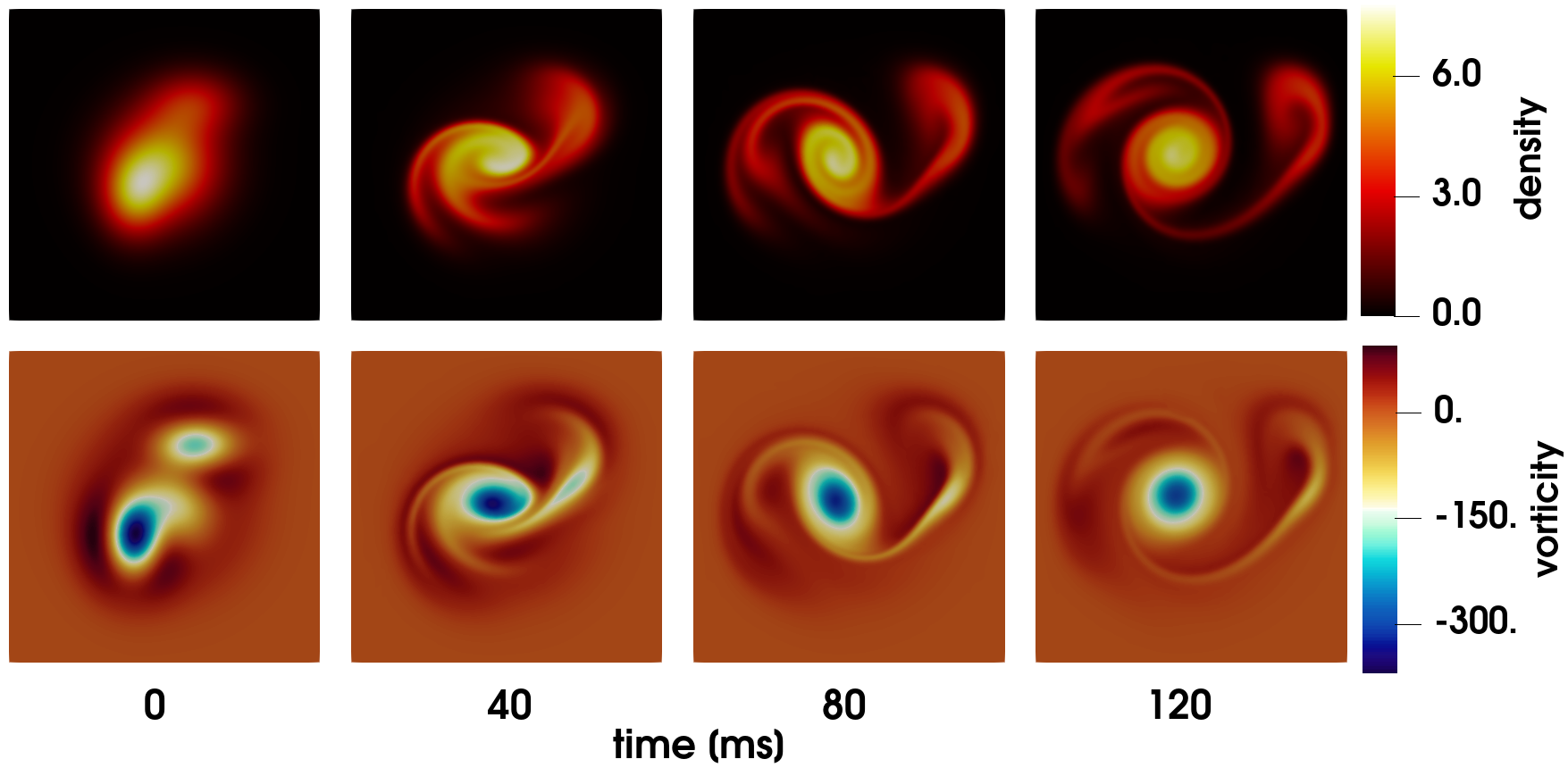}
  \caption{Simulation of the Hasegawa-Wakatani model in a 2D slab initialised with two neighbouring blobs. A full evolution animation is available on Zenodo \cite{Zenodo-Movies}.}
  \label{FIGURE_HW_galaxies}
\end{figure}

Figure-\ref{FIGURE_HW_galaxies} shows the evolution of the dynamics for the density and vorticity variables. This simulation requires approximately 1s per epoch,  which corresponds to about 5min per time-step with 300 epoch per time-step. Figure-\ref{Figure_intro}b shows a similar simulation with multiple initial Gaussians leading to turbulent mixing, demonstrated with 25 FEM-NN nodes (with 2251 parameters per FEM-NN node per variable). That simulation requires approximately 1.5s per epoch, which corresponds to about 7.5min per time-step with 300 epoch per time-step.

Although these simulations may be feasible with a smaller number of FEM-NN nodes, running such cases on a single PINN network per variable is challenging. With ENGD, a single NN for the whole domain is restricted by memory usage due to the high impact of the auto-differentiation graph required for ENGD's model parameter derivatives (which are computed for each variable and their auto-diff spatial derivatives). This limits the maximum NN sizes. Larger NN models are possible with Adam or L-BFGS, but these are too slow to converge to appropriate accuracies.

In terms of computational efficiency, 5 minutes per time step with 13 FEM-NN nodes (hence 39 GPUs in total) for the Figure-\ref{FIGURE_HW_galaxies} case corresponds to approximately 0.8 node-hours (4 Nvidia H100 GPUs per compute node). In comparison, the same JOREK simulation requires 1.15s per time-step on 2 compute nodes (256 AMD CEPYC 9745 CPU cores per node), which amounts to approximately 0.6$\times10^{-3}$ node-hours. In terms of wall-clock time per step, this is approximately 260 times longer, which is encouraging, but in terms of compute cost, this is a gap of approximately 3 orders of magnitude. PirateNets do not report direct comparisons with conventional solvers, but they do report approximately 3 hours per time-window for the 2D Kolmogorov flow simulations \cite{Wang2025}, which is coherent with our 0.8 node-hours (= 3.2 GPU-hours) per time-step.

\begin{figure}[ht!]
  \centering
        \includegraphics[width=0.5\linewidth]{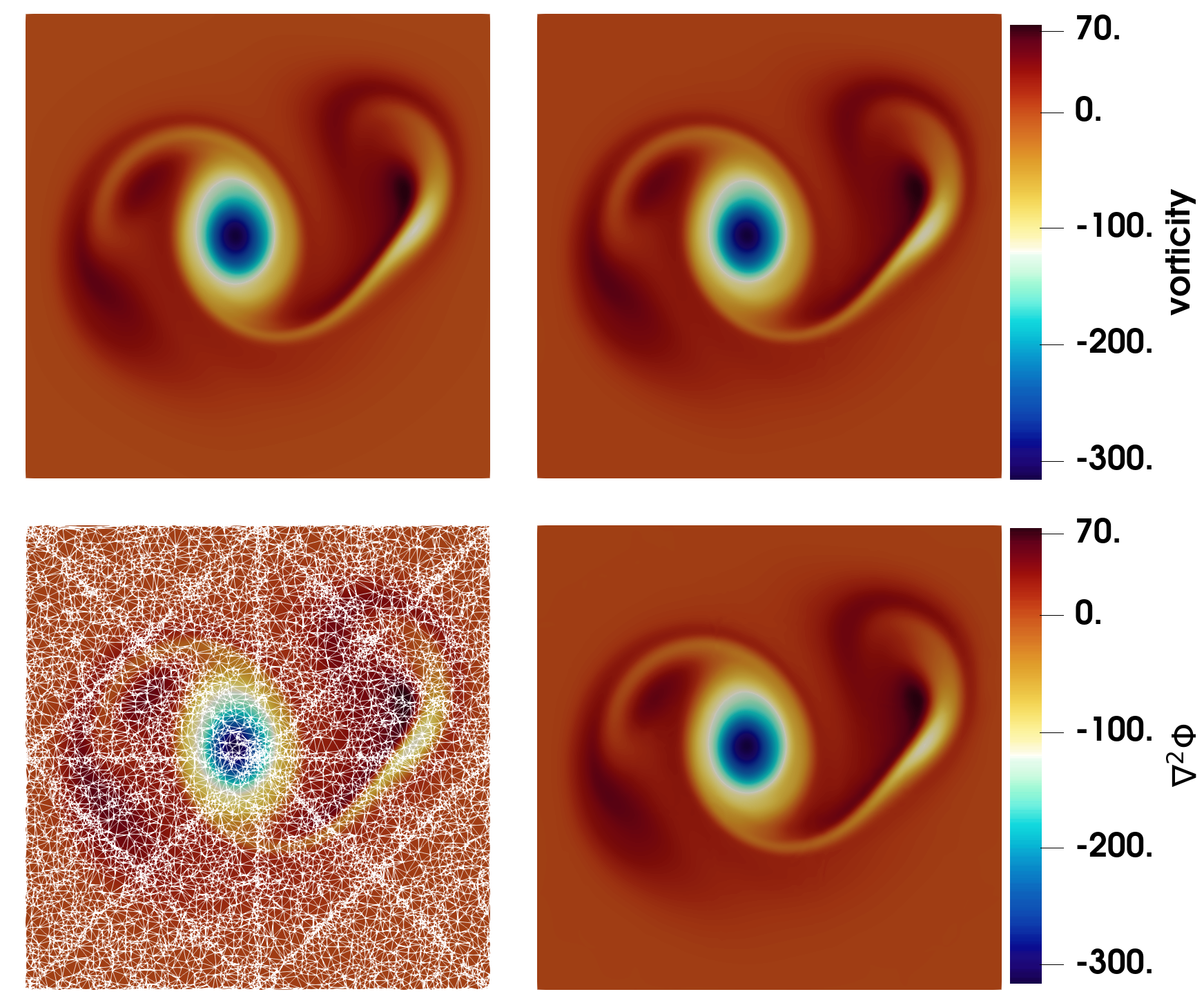}
  \caption{Comparison of the Poisson solve within the Hasegawa-Wakatani model, at time-step $n_t=80$. Top left: Vorticity field from a \cite{JOREK} simulation with identical conditions, showing agreement with Fuse-PINN. Top right: Vorticity field from Fuse-PINN. Bottom right: $\nabla^2\Phi$ field from Fuse-PINN showing good agreement with vorticity field and smoothness of derivatives. Bottom left: FEM-NN elements and nodes for the Fuse-PINN simulation (with lower sampling for visibility).}
  \label{FIGURE_HW_poisson}
\end{figure}

One of the key aspects of evolving models like Hasegawa-Wakatani is the need for a robust solve of the Poisson (vorticity) equation-\ref{HW_poisson}. Figure-\ref{FIGURE_HW_poisson} shows the comparison of the vorticity field and the Laplacian $\nabla^2\Phi$ after 80 time-steps. The Laplacian gradients in the figure are calculated with the auto-differentiation, exactly as in the training residual calculation, and clearly show the smoothness of second order derivatives across FEM-NN element edges. Comparison against the classical numerical solver JOREK also shows reasonable agreement.

\subsection{Electrostatic Reduced-MHD simulations}

The second advanced, time-evolutive, non-linear model implemented in Fuse-PINN is the electrostatic Reduced-MHD model, as described in \cite{Pamela2025}. It is similar to the Hasegawa-Wakatani model in the sense that it is also a reduced, electrostatic, fluid turbulence model in the presence of a strong magnetic field. The main differences are that it evolves both density and temperature, and that the projection in poloidal plane for the momentum equation and the velocity ansatz are slightly different.

The electrostatic Reduced-MHD model solves for 3 physics variables, the density $\rho$, the temperature $T$ and the electric potential $\Phi$. As in the Hasegawa-Wakatani implementation, an auxiliary variable $w$ is used to reduce the order of direct derivatives needed in the residual formulation. The equations are described in details in Appendix-\ref{AppendixB_ERMHD}.
\begin{figure}[ht!]
  \centering
        \includegraphics[width=0.98\linewidth]{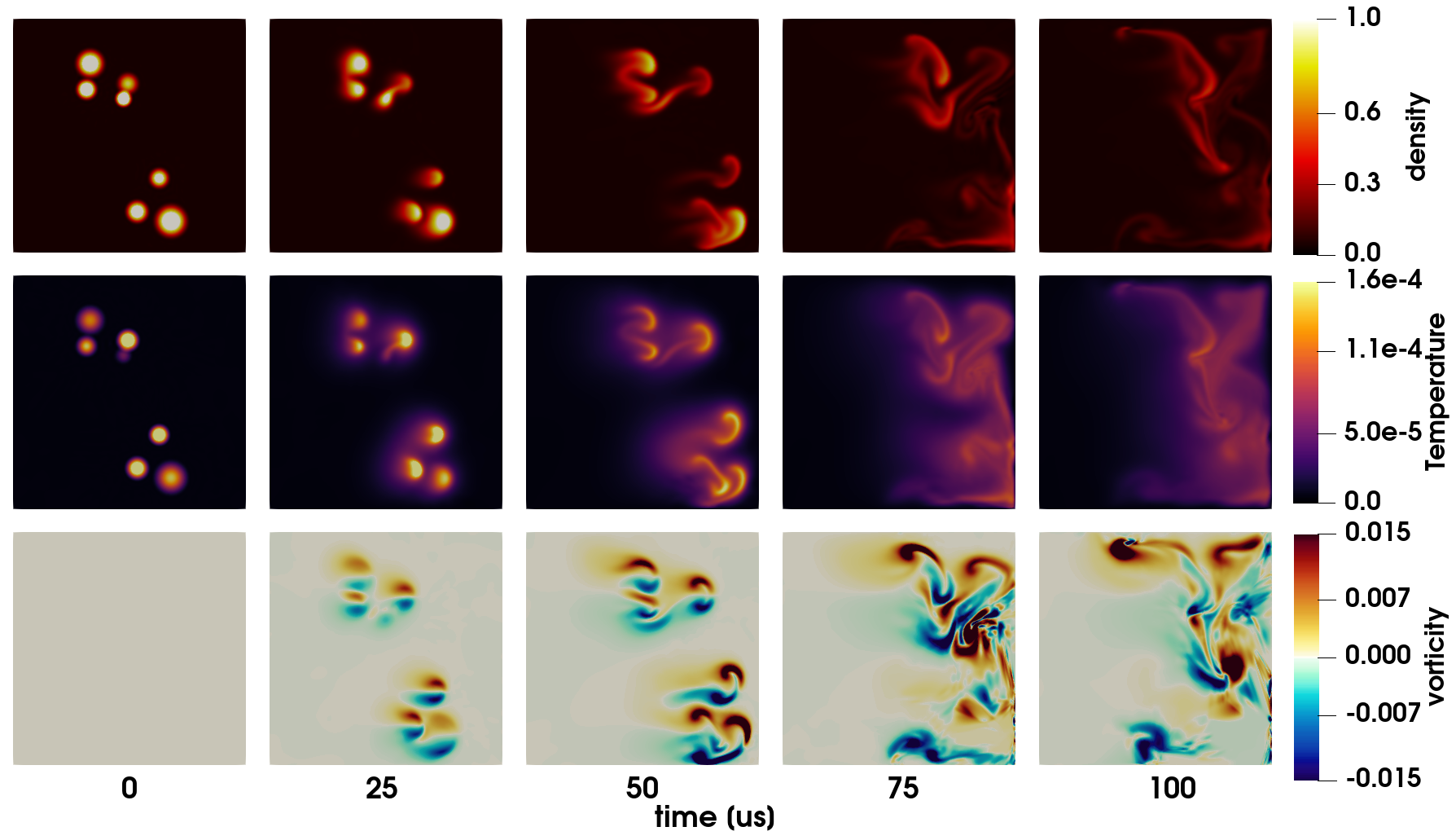}
  \caption{Simulation of the electrostatic Reduced-MHD model with blobs evolving radially in toroidal geometry. A full evolution animation is available on Zenodo \cite{Zenodo-Movies}.}
  \label{FIGURE_ES-RMHD_blobs}
\end{figure}

The simulation presented here reproduces the JOREK simulation of Figure-4 in \cite{Pamela2025}, using the same viscosity, diffusion and initial conditions. It is run in a 2D square domain of $x\in[10,11]$ and $y\in[-0.5,0.5]$, using a resolution of 25 FEM-NN nodes, with a total sampling of 640$\times$640. The simulation is evolved for 140 time-step of $0.75\times10^{-6}s$. Each FEM-NN node evolves a NN of 1051 parameters for each variable. Figure-\ref{FIGURE_ES-RMHD_blobs} shows the evolution of the dynamics for the density, temperature and vorticity variables. While the initial radial evolution of the blobs reproduces the original JOREK case, once the blobs reach the outer wall, precision drops and numerical noise becomes visible in the vorticity. Better precision might be achieved with other implementations of boundary conditions, or possibly with smaller time-steps. Nevertheless, this simulation shows the ability of the Fuse-PINN implementation to retain numerical stability over long simulations with complex convection dynamics.

\subsection{Scalings}\label{Section_scalings}

One of the key advantages of the parallelisation method implemented in Fuse-PINN is that it scales linearly with the size of the problem $N_p$, the total number of degrees of freedom (or in our case the total number of trainable parameters), whereas conventional sparse matrix solvers scale as $O(N_p^{1.5})$ for 2D problems, and even $O(N_p^2)$ for 3D problems.

\begin{figure}[ht!]
  \centering
  \resizebox{1.0\linewidth}{!}{
    \begin{floatrow}
      \ffigbox{
        \includegraphics[width=0.90\linewidth]{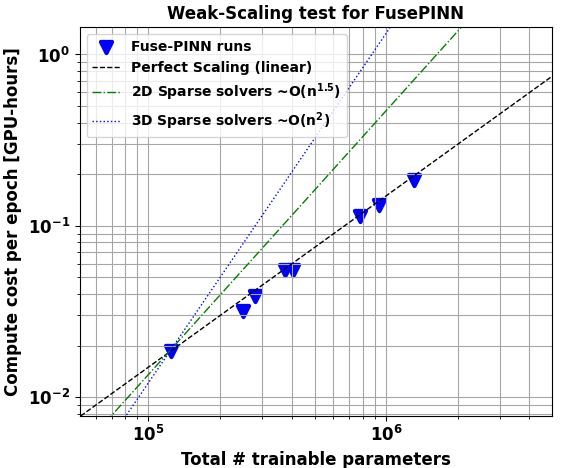}
      }
      {\caption{Weak scaling test using Nvidia H100 GPUs, showing near-perfect linear scaling.}
        \label{FIGURE_scalings}
      }
        \capbtabbox{
          \begin{tabular}{|m{0.2\linewidth}|m{0.21\linewidth}|m{0.17\linewidth}|m{0.15\linewidth}|} \hline
          \footnotesize
          FEM-NN nodes & Total \# of parameters & Time per epoch & Comm. time \\ \hline
          \hline
          4  & \textit{ }\textit{ }\textit{ }124,812 & 5.6s & 1.0e-2s \\
          \hline
          8  & \textit{ }\textit{ }\textit{ }249,624 & 4.8s & 1.1e-2s \\
          \hline
          9  & \textit{ }\textit{ }\textit{ }280,827 & 5.2s & 1.0e-2s \\
          \hline
          12 & \textit{ }\textit{ }\textit{ }374,436 & 5.5s & 1.5e-2s \\
          \hline
          13 & \textit{ }\textit{ }\textit{ }405,639 & 5.1s & 1.4e-2s \\
          \hline
          25 & \textit{ }\textit{ }\textit{ }780,075 & 5.5s & 1.7e-2s \\
          \hline
          30 & \textit{ }\textit{ \textit{ }}936,090 & 5.3s & 1.6e-2s \\
          \hline
          42 &                             1,310,526 & 5.5s & 1.5e-2s \\
          \hline
          \end{tabular}
        }
        {\caption{Computing times per epoch for the scaling, including communications.}
          \label{Table-scaling-timings}
        }
    \end{floatrow}
  }
\end{figure}

A Weak scaling test is executed for Fuse-PINN using the Hasegawa-Wakatani simulation of Figure-\ref{Figure_intro}. In this test, the size of the domain (the total number of collocation points) is increased linearly with the number of FEM-NN nodes (and hence the number of GPUs), while the size of the NN model on each FEM-NN node is kept constant. A relatively large NN model is used for this test, with 10401 parameters, resulting in a total problem size of $\sim$$1.3M$ parameters and a 548x548 grid for the largest case. For reference, the largest PirateNets models used for 3D Taylor-Green Vortex simulations was $\sim$$4.7M$ parameters \cite{Wang2025}). The results are shown in Figure-\ref{FIGURE_scalings}, demonstrating near-perfect linear scaling; the $O(n^{1.5})$ and $O(n^2)$ scalings are shown for comparison.

A rigorous strong scaling is not yet possible as the implementation currently ties every FEM-NN node to its own GPU. Multiple FEM-NN nodes per GPU would allow a strong scaling, but this was not the priority so far as it effectively goes counter to our ambition of massively parallel capability. Such options will be explored in future work. However, given the negligible communication times shown in Table-\ref{Table-scaling-timings}, there is no reason to suggest a strong-scaling test wouldn't give a near-perfect scaling like the weak-scaling test.

Nevertheless, the scaling test presented here considers timing per epoch, which is not exactly a full time-step solve, for which many epochs are required. A complete convergence study with rigorous precision levels comparison against conventional solvers is beyond the scope of this initial achievement. In fact, before addressing such detailed scaling comparisons, the overall performance of the implementation needs to be improved, as it is currently not competitive with conventional solvers in the first place, by approximately a factor 1,000. Performance optimisation of the implementation is discussed in the Conclusion below.


\section{Conclusion}

\subsection{Summary}

This study presented Fuse-PINN, a scalable framework that combines Physics-Informed Neural Networks with a finite-element-inspired domain decomposition, an implicit time integration, and an Energy Natural Gradient Descent optimiser for high precision. The FEM-NN formulation distributes separate neural networks across mesh nodes while preserving smooth solutions and derivatives across arbitrary triangular element boundaries. This enables nonlinear, multi-field PDE systems to be distributed across multi-node, multi-GPU architectures, with predominantly local communications. Fuse-PINN was demonstrated on fusion-relevant physics models like Hasegawa-Wakatani and electrostatic reduced-MHD, including convection-dominated dynamics and implicit Poisson solves. The implementation achieves near-perfect linear scaling on weak scaling tests, as the problem size and number of GPUs are increased, with communications representing only a negligible fraction of the total compute cost.

These results demonstrate that PINN-based methods can be formulated as genuinely scalable scientific-computing workloads and are not restricted to small, single-device problems. Nevertheless, the present work should be regarded as a proof of concept for scalability and numerical capability, rather than a performance-competitive replacement of mature conventional solvers, for which further development and optimisation are needed.

\subsection{Limitations and Future Work}

The principal limitation of Fuse-PINN is its current computational cost. By comparison, the results presented here indicate an overall computational efficiency gap of approximately three orders of magnitude relative to a conventional solver. Although a rigorous comparison using identical equations, resolution, convergence tolerances, hardware allocation, and accuracy metrics remains necessary, it is evident that significant optimisation improvements are necessary for a production-ready framework. Nevertheless, Fuse-PINN seems to have performance costs comparable to other large-scale PINN models like PirateNets, which is encouraging.

A substantial part of this performance gap is attributable to implementation maturity rather than to the FEM-NN parallelisation principle itself. For example, the integration of the physics residual across FEM-NN neighbour nodes and parent elements is still executed through explicit Python-level loops, which could be vectorised. Likewise, there may be faster ways of implementing the parameter-differentiation required by ENGD: this is currently done directly on the physics variables, which are then assembled into the equation residuals; whereas differentiating the equations directly (after assembly) may be more efficient. Although PyTorch \texttt{vmap} is already used in constructing the neural-network and spatial derivatives required for the Gram matrix, the resulting differentiation graph remains computationally and memory intensive, and PyTorch compilation only supports single-derivative levels, which is not sufficient for our implementation. Therefore, a re-implementation of the framework in JAX should be seriously considered as compilation alone would certainly improve performance, however that also comes with challenges in terms of parallelisation due to the nature of the FEM-NN formulation, which would need to be formulated in functional form.

Future work must also include controlled convergence and accuracy studies against other established solvers, with strong-scaling tests enabled by allocating multiple FEM-NN nodes to each GPU, and systematic profiling of complete time steps. Of course, the framework needs to be extended to 3D meshes. Additional priorities include stronger boundary conditions, more robust positivity-preserving formulations, as well as adaptive sampling and heterogeneous allocation of neural-network sizes for $h^p$-refinement. Collectively, these developments are required to determine whether Fuse-PINN’s encouraging initial scaling can translate into a competitive time-to-solution avenue for large, nonlinear fusion simulations.


\subsubsection*{Acknowledgments}

This work has been carried out within the framework of the EUROfusion Consortium, funded by the European Union via the Euratom Research and Training Programme (Grant Agreement No 101052200 — EUROfusion). Views and opinions expressed are however those of the author(s) only and do not necessarily reflect those of the European Union or the European Commission. Neither the European Union nor the European Commission can be held responsible for them. This work has been carried out within the framework of the RCUK Energy Programme [grant number EP/I501045]. This work was performed using the Pitagora computer at CINECA in Italy, within the EUROfusion framework. This work was performed with the support of the JOREK Team [See https://www.jorek.eu for the present list of team members].

\bibliographystyle{unsrt}
\bibliography{my_references}


\newpage

\appendix
\section{Appendix: Detailed Methods}\label{AppendixA}

\subsection{Time-discretisation}

Fuse-PINN uses the so-called discrete time-integration. As introduced in \cite{Raissi2019}, instead of optimising a single Neural Network to learn the solution of an entire simulation trajectory, the optimisation is done one time-step at a time. This allows the initialisation of the next time-step with the previous NN solution, greatly accelerating convergence. It also removes the balancing issue found when initial condition losses are competing with domain losses. Additionally, it separates concerns between the solver method and the time-duration of the simulation, allowing for extended time-integration, as well as easier check-pointing and restart. Lastly, it allows for much smaller Neural Networks, which is essential when scaling towards large use cases.

At present, the fully implicit Crank-Nicolson time-stepping method is implemented in Fuse-PINN. Let $\mathbf{u}\in\mathbb{R}^m$ be the solution of a PDE system with $m$ equations defined in a spatial domain $\mathbf{X}\in\mathcal{D}\in\mathbb{R}^d$ inside the boundary domain $\mathcal{B}\in\mathbb{R}^{d-1}$. The evolutive formulation of the PDE system, assuming it involves only up to first-order time-derivatives, is
\ba
  \frac{\partial\mathbf{u}}{\partial t} &=& F(\mathbf{u}) \ttt{ }\ttt{ } \textnormal{in}\ttt{ }\ttt{ } \mathcal{D}, \\
  G(\mathbf{u}) &=& g\ttt{ }\ttt{ } \textnormal{on}\ttt{ }\ttt{ } \mathcal{B}.
\ea
With the discretised solution $\mathbf{u}_n$ at a given time $t_n$, the Crank-Nicolson method provides a formulation at the next time-point $t_{n+1}$ as
\be
  \mathbf{u}_{n+1} = \mathbf{u}_{n} + \frac{\Delta t}{2} \left[F(\mathbf{u}_{n+1}) + F(\mathbf{u}_{n})\right]. \label{Crank-Nicolson}
\ee

Implicit time-integration is essential for non-linear PDEs, particularly in fusion applications, where time-scales and spatial-scales can have orders of magnitude variations in a given physics model, such as fast-waves in full-MHD models, or anisotropic diffusion for transport along/across magnetic field lines. Implicit schemes allow much larger time-step sizes that are not bound to the grid resolution like explicit schemes (CFL condition).

The main challenge of implicit methods is that, due to the $F(\mathbf{u}_{n+1})$ component on the RHS of equation-(\ref{Crank-Nicolson}), they typically result in a large linearised system, meaning a large matrix needs to be inverted. This typically restricts the scalability of classical approaches due to the memory limits on current HPC hardware, leading to inefficient parallelisation scaling at large resolutions. This paper shows how Fuse-PINN provides a promising avenue with a compute cost which scales linearly with resolution, whereas classical sparse matrix solvers typically scale as $O(n^{1.5})$ or $O(n.log(n))$.

The generalisation of the Fuse-PINN implementation should be relatively straight-forward for other methods such as Gear's or Euler, but is left for future work as Crank-Nicolson is sufficient for the proof-of-concept addressed here.

\subsection{Separation of variables}

Most PINN implementations use a single Neural Network for all variables at once. As in \cite{He-Nature2024,Laubscher2021}, having explored both options, we found that competing variables within a given equation loss can significantly degrade, even prevent, the optimisation of a single Neural Network. In line with our ambition to aim at large-scale HPC applications and parallelisation of the method, a segregated PINN approach is adopted.

Beyond parallelisation and optimisation reliability, a segregated approach has other advantages, particularly in terms of the Energy Natural Gradient Descent, where the equations need to be linearised with respect to the trained model, which will be discussed below.

\subsection{Finite-Element Method}

Fuse-PINN uses a FEM approach with triangular elements. In typical FEM implementation from classical numerical methods, degrees of freedoms (DOFs) at the element nodes are combined with basis functions across the element's dimensions to obtain a solution on the element domain. In Fuse-PINN, we use a single basis function at each node, with a common overlap region in the middle of the element, to combine the NN solutions from the three element nodes. In other words, this approach can be described as a single non-scalar, non-linear DOF with a single basis function.

One of the great advantages of PINNs often highlighted is their mesh-free ability to describe a solution on a randomly sampled domain. However, in order to scale towards complex non-linear PDEs in large spatial domains, a spatial decomposition of some kind is necessary. The present work is inspired by some pioneering work by several teams in domain decomposition. The key challenge in domain decomposition is that information needs to be communicated between sub-domains through so-called overlap regions.

The notable work by \cite{Moseley2021} introduced a FEM-like approach where the domain is decomposed in rectangular patches, enabling spatially reliable overlapping regions. This team also explored mixed methods like TensorGalerkin where a tensor formulation of the FEMs allows for a classical solver and a PINN to be used on the same framework \cite{Wen2026}. Another impressive work on domain decomposition is certainly that of Shukla et al. \cite{Shukla2021}, where arbitrary sub-domains are defined with complex boundaries, with successful parallelisation. Other similar works explored variations of these approaches \cite{Ye_ARC2025,Huang2025}.

So why not simply implement one of these approaches? Despite its success, the FB-PINN method is restricted by the regular nature of the rectangular decomposition, making it difficult to generalise to arbitrary domains. The decomposition does not necessarily need to be rectangular, but one of the complexity in implementing a generic decomposition with overlap regions is how to deal with multiple overlaps from several neighbour sub-domains when these overlap regions are of random geometry. Having a triangular decomposition with a strictly local representation enables a fully generic approach which can be adapted to any domain. Aside from generality, the overlap regions between sub-domains are defined as a discrete region -- a point is either inside or outside the overlap layer -- which cannot guarantee smoothness of the solution's derivatives across the overlap's boundaries, and effectively introduces a singularity that the NN models need to compensate for. The approach by FB-PINN and Huang et al. resolves this issue by introducing either smooth $C^{\infty}$ partition-of-unity or ``s-shaped'' weight functions at the overlap regions \cite{Moseley2021,Huang2025}. However both approaches are restricted to weight functions that are defined on regular rectangular domain decomposition, which is challenging to generalise to arbitrary geometries. The cPINN/xPINN approach by Shukla et al. \cite{Shukla2021} stitches sub-domains at a boundary, without any overlap, transferring information between sub-domains with gradient updates during training, but it is unclear how efficient this method would be for solving global problems like Poisson equations.

The Fuse-PINN approach is inspired by these pioneering advances to incorporate the best of their advantages into a generalised method. It uses triangular sub-domain regions with smooth overlap weight functions within each triangle, formulated in a way which guarantees continuity of derivatives across element edges. Achieving this smoothness was one of the key challenges of this work, which is discussed in details below.

For the purpose of clarity, it is helpful to define a clear indexing convention for the rest of the paper. The architecture of the Fuse-PINN implementation splits the PINN training in three dimensions: the FEM-NN grid nodes, the physics variables, and the time-steps. Therefore, the Neural-Networks trained for the solution are indexed as
\ba
  \mathcal{N}_{i,k,n} \ttt{ }\ttt{ }\textnormal with: && \label{NN_indexing}\\
  i &\rightarrow& \textnormal{index of a FEM-NN node on the grid.} \nonumber\\
  k &\rightarrow& \textnormal{index of the physics variable (and its corresponding equation) being trained.} \nonumber\\
  n &\rightarrow& \textnormal{index of the current time-step, ($n$+1) being the next time-step being solved.} \nonumber
\ea
where a given Neural Network solution is valid on all $n_p$ parent elements of a given node, which are typically indexed with $0\le j < n_p$. Similarly, the physics variables (the components of $\mathbf{u}_{j,n}$)  are defined per element and indexed as
\ba
  u_{j,k,n} \ttt{ }\ttt{ }\textnormal with: && \label{variable_indexing}\\
  j &\rightarrow& \textnormal{index of a FEM-NN element on the grid.} \nonumber
\ea
These are the components of the array $\mathbf{u}_{j,n}$ which represents the full solution $\mathbf{u}_{n}$ on a given element $j$, and defined as follows.

\subsubsection{Triangular elements}

Let a triangular element in 2D have corner nodes $(\mathbf{C}_1,\mathbf{C}_2,\mathbf{C}_3)$, each given by their cartesian coordinates $(x_i,y_i)$, with $1\le i\le3$. Let this element have index $j$ in the FEM-NN grid. The triangle's barycentric coordinates $(r,s,t)$ can be used to locate any point $(x,y)$ inside the element, such that
\ba
  x &=& r\ttt{ }x_1 \ttt{ }+\ttt{ } s\ttt{ }x_2 \ttt{ }+\ttt{ } t\ttt{ }x_3, \\
  y &=& r\ttt{ }y_1 \ttt{ }+\ttt{ } s\ttt{ }y_2 \ttt{ }+\ttt{ } t\ttt{ }y_3, \\
  &&r + s + t = 1.
\ea
Similarly, at any given point $(x,y)$ and time $t_n$, each $u_{j,k,n}$ scalar field component of $\mathbf{u}_{j,n}$ on this element $j$ can be expressed as the sum of three nodal Neural Networks at the triangle's corners:
\ba
  u_{j,k,n} &=& w_1\mathcal{N}_{1,k,n} + w_2\mathcal{N}_{2,k,n} + w_3\mathcal{N}_{3,k,n}, \label{nodal_formulation}\\
  &&w_1 + w_2 + w_3 = 1,
\ea
where $w_i(x,y)$ are the weight functions which ensure a smooth overlap between the contributions of each node's NN. Note that, for simplicity, we have indexed the three NN's with $1\le i\le3$, but in a FEM-NN grid with more than 3 nodes, the exact form would be
\ba
  u_{j,k,n} &=& w_1\mathcal{N}_{i_1,k,n} + w_2\mathcal{N}_{i_2,k,n} + w_3\mathcal{N}_{i_3,k,n}, \\
  &&w_1 + w_2 + w_3 = 1,
\ea
where $i_1,i_2,i_3$ are the corresponding indices from the full node list. For clarity (and because this distinction is not strictly necessary in the remainder of this section), let us keep the simpler indexing and reformulate definition-(\ref{nodal_formulation}) as
\ba
  u_{j,k,n} &=& \sum_{i=1}^{3} w_i\mathcal{N}_{i,k,n} , \label{nodal_formulation2}\\
  \sum_{i=1}^{3} w_i &=& 1.
\ea

\subsubsection{Overlap weight functions}\label{Section-Weight-Functions}

The careful geometrical choice of these basis functions, or weight functions, is critical for a successful implementation that can generalise to any triangular mesh. One of the main reasons for this can be quickly understood by considering the spatial derivatives of a given scalar $u_{j,k,n}$, given by applying the chain-rule to equation-(\ref{nodal_formulation2}), for example $\partial_x u_{j,k,n}$,
\be
  \frac{\partial u_{j,k,n}}{\partial x} =
    \sum_{i=1}^{3} \left( \frac{\partial w_i}{\partial x}\mathcal{N}_{i,k,n}
    + w_i\frac{\partial\mathcal{N}_{i,k,n}}{\partial x} \right),\label{weight_derivatives}
\ee
where the derivatives $\partial_x w_i$ simply depend on the formulation we prescribe, and where the NN derivatives $\partial_x\mathcal{N}_{i,k,n}$ are obtained through auto-differentiation. Hence, equation-(\ref{weight_derivatives}) implies that any discontinuity of the weight function derivatives $\partial_x w_i$ across elements will result in a discontinuity of the full derivative $\partial_x u_{j,k,n}$. Likewise for higher-order derivatives, and for $y$-derivatives. In other words, it is important that the weight functions are defined in a way that ensures their $C^n$-continuity across elements at least up to the same order of spatial derivatives involved in the PDE system.

Naturally, each weight function should be 1 at the corresponding node, and decay to 0 away from the node (ideally about halfway through the element). All weights need to have some overlap around the centre of the element, and each pair of weight functions need to overlap approximately around the middle of their common edge. Therefore, an easy way to construct these functions is to use the medians of the triangle, which can be expressed as functions of the barycentric coordinates. For example, the median through node $\mathbf{C}_1$, where $r=1$, $s=t=0$, is given by
\be
  s = t = \frac{1 - r}{2}, \ttt{ }\forall\ttt{ }r\in[0,1].
\ee
Weight functions defined with a $tanh()$ located at the two medians through the opposite nodes is already a practical solution. Unfortunately there is no guarantee that the medians on either side of a common edge between two triangles will be aligned (parallel), and the resulting angle will lead to discontinuous derivatives of $w_i$.

It is possible to use a mixture of the medians and the binormals at each edge, with a $tanh()$ that depends on the binormals at the edges and on the medians as we get further into the element. However, while this solution is ``less discontinuous'' than using just the medians, it still does not guarantee continuity of all derivatives. This is because exact (mathematical) continuity would require a formulation based not just on the barycentric coordinates $(r,s,t)$, but on an affine transformation of the barycentric coordinates based on the neighbour triangles. While possible, this solution leads to weight functions that depend on the geometry of neighbour elements, which is inconvenient. A strictly local definition is far more preferable (and easier to implement).

\begin{figure}[ht!]
  \centering
        \subfigure[ ]{\includegraphics[width=0.49\linewidth]{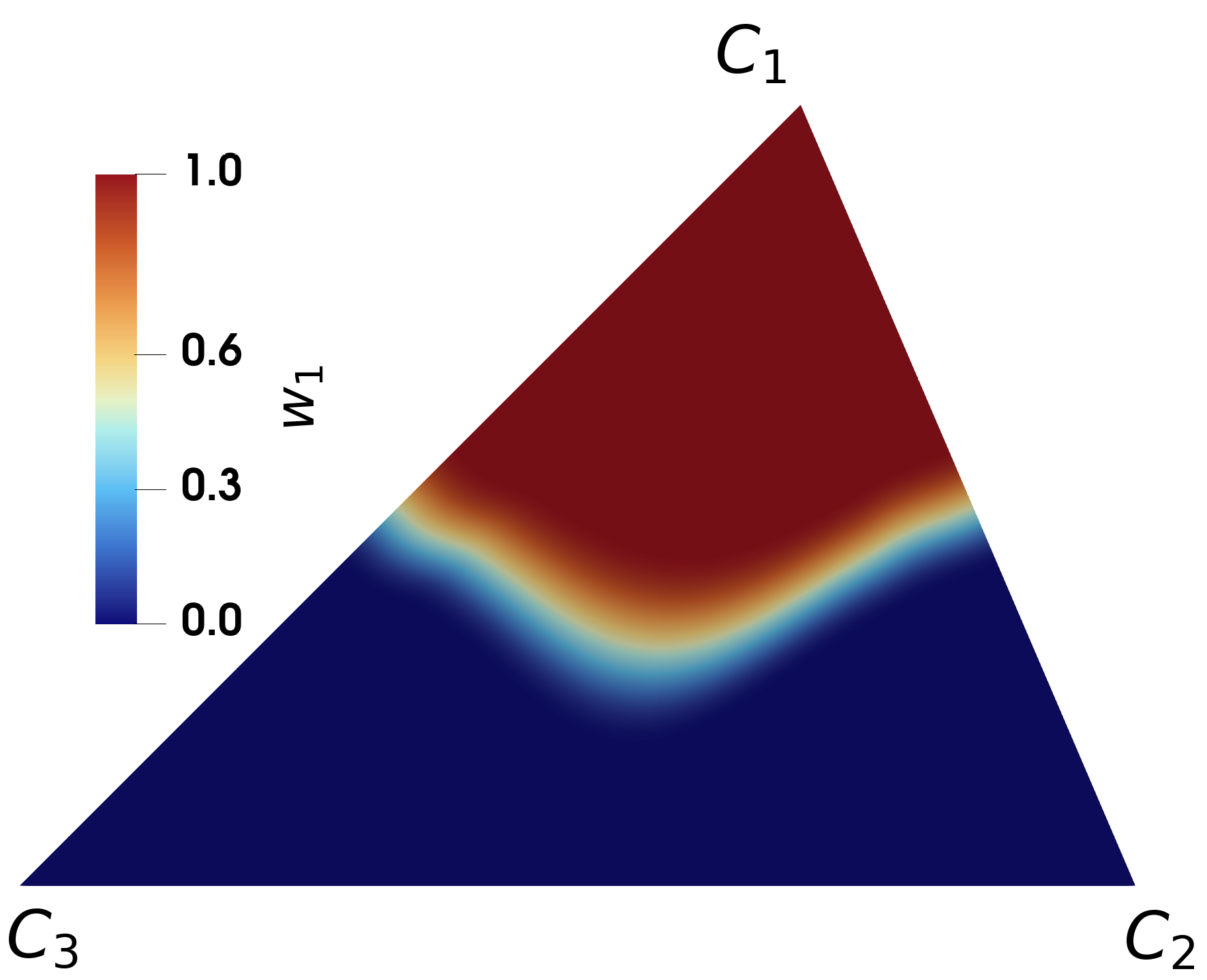}}
        \subfigure[ ]{\includegraphics[width=0.49\linewidth]{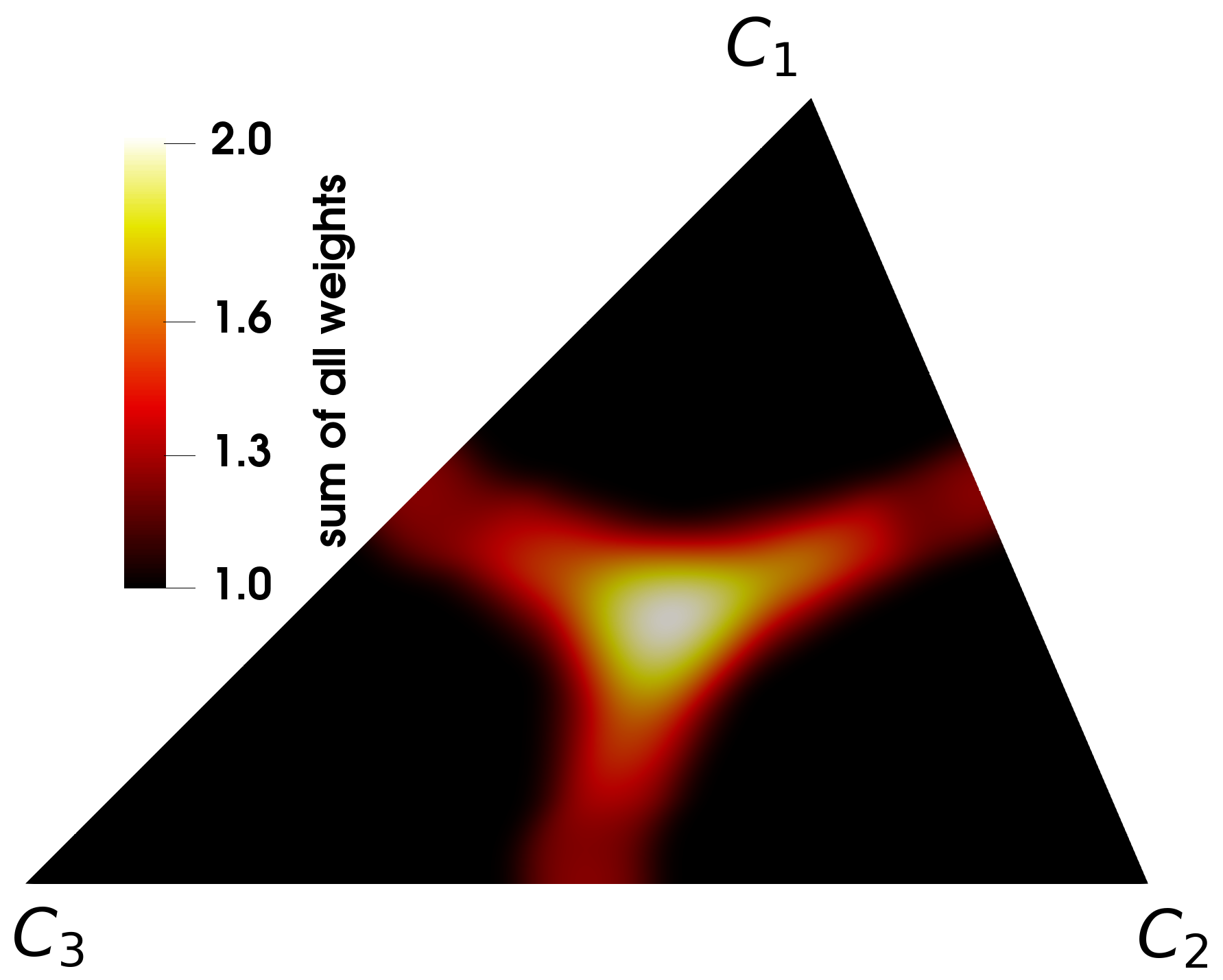}}
  \caption{A single triangular element showing \textnormal{(a)} the weight of the $C_1$ node, $w_1(x,y)$, \textnormal{(b)} the sum of all weights (before normalisation), illustrating the overlap regions where the solution is shared between the node NN contributions. Where the sum is highest and reaches $\sim$2, all three node NN's are contributing, but where the sum is lower, $\sim$1.5, only two node NN's are contributing.}
  \label{Figure_weight_functions}
\end{figure}

Instead, the weight functions are defined to depend on the medians at the centre of the elements, transitioning to the spatial distance of the mid-point at each edge. Since the spatial distance is guaranteed to be the same on either side of a triangle edge, continuity of $w_i$ derivatives is achieved. The exact formulation for $w_1$ is defined, at any point $\mathbf{X}=(x,y)$ with barycentric coordinates $(r,s,t)$, as follows. Let
\ba
  \mathbf{X}_{L} &=& \frac{1}{2}(\mathbf{C}_1+\mathbf{C}_2),\\
  \mathbf{X}_{R} &=& \frac{1}{2}(\mathbf{C}_1+\mathbf{C}_3),
\ea
be the midpoints along the (left and right) edges of corner $C_1$. Let
\ba
  \mathbf{M}_{L} &=& \frac{1 - t}{2}\mathbf{C}_1 + \frac{1 - t}{2}\mathbf{C}_2 + t\mathbf{C}_3, \\
  \mathbf{M}_{R} &=& \frac{1 - s}{2}\mathbf{C}_1 + s\mathbf{C}_2 + \frac{1 - s}{2}\mathbf{C}_3, 
\ea
be the median lines across the edges of $C_1$: through $C_3$ on the left, and through $C_2$ on the right. Then the distances from $C_1$ to those points can be defined as
\ba
  D_{1L} &=& |\mathbf{C}_1 - \mathbf{X}_{L}|,\\
  D_{2L} &=& |\mathbf{C}_1 - \mathbf{M}_{L}|,\\
  D_{1R} &=& |\mathbf{C}_1 - \mathbf{X}_{R}|,\\
  D_{2R} &=& |\mathbf{C}_1 - \mathbf{M}_{R}|,
\ea
This enables us to define a decay location for the weight which transitions smoothly from $D_{1L}$ to $D_{2L}$ to $D_{2R}$ to $D_{1R}$, which can be formulated as
\ba
  D_{ref} &=& D_L + T_1 (D_R - D_L),\\
  D_L &=& D_{1L} + T_L (D_{2L} - D_{1L}),\\
  D_R &=& D_{1R} + T_R (D_{2R} - D_{1R}),
\ea
where $T_1$, $T_L$ and $T_R$ are smooth transition functions from 0 to 1. Given this reference distance $D_{ref}$, the decay width of the weight now needs to be adjusted between the left and right sides. This is important in case one of the sides is much larger than the other. This is formulated as
\be
  \delta = \delta_L + T_2 (\delta_R - \delta_L),
\ee
where $T_2$ is another transition function. The widths on the left and right are simply proportional to the side lengths, with a user-input parameter $\lambda$, such that $\delta_L = \lambda D_{1L}$ and $\delta_R = \lambda D_{1R}$. Finally, given the distance $D = |\mathbf{X} - \mathbf{C1}|$, the weight function is defined as
\ba
  w_1 &=& 1 - \mathcal{T}(D, \sigma D_{ref}, \delta),
\ea
where $\mathcal{T}$ is a so-called $C^{\infty}$ bump-based smooth step function, similar to $tanh()$, but which goes exactly to 0 (and 1) below (or above) the transition width. It is essential that a decay or transition function becomes exactly 1 or exactly 0 beyond the decay width, otherwise even tiny non-zero weight values can result in a NN training trying to compensate for the imbalance with huge values. This pseudo-tanh function is defined for a given distance $d$, position $p$ and width $w$, as
\ba
\mathcal{T}(d, p, w) &=& \frac{E_1}{E_1 + E_2}, \\
  E_1 &=& exp\left(\frac{-1}{d_n + \epsilon}\right), \\
  E_2 &=& exp\left(\frac{-1}{1 - d_n + \epsilon}\right), \\
  d_n &=& \frac{d - (p-w)}{2w}, \nonumber
\ea
where $\epsilon$ is a small parameter to account for machine precision, typically $\sim$$10^{-12}$. The transition functions mentioned previously are also defined with the same custom step function,
\ba
  T_1 &=& \mathcal{T}\left(t, \frac{1 - r}{2}, \delta_1\right),\\
  T_2 &=& \mathcal{T}\left(t, \frac{1 - r}{2}, \delta_2\right),\\
  T_L &=& \mathcal{T}\left(3t, \sigma_3, \delta_3\right),\\
  T_R &=& \mathcal{T}\left(3s, \sigma_3, \delta_3\right),
\ea
where the factor 3 in $T_L$ and $T_R$ is because the centre of the triangle in barycentric coordinates is at $(r,s,t) = \left(\tfrac{1}{3},\tfrac{1}{3},\tfrac{1}{3}\right)$.

A similar formulation can be repeated for $w_2$ and $w_3$, and all three weight functions are of course normalised, e.g. $w_1 = w_1 / (w_1+w_2+w_3)$. The derivatives of the $w_i$ weights, needed for the chain-rule equation-(\ref{weight_derivatives}) are obtained using auto-differentiation. This leads to a number of control parameters to adjust the widths and positions of all the control functions, which are listed in Table-\ref{Table_weight_functions}.
\begin{table}[h!]
\centering
\begin{tabular}{ |c|c|c| } 
 \hline
 \multicolumn{3}{|c|}{\textbf{Weight Functions Control Parameters}} \\
 \hline
 \textbf{parameter} & \textbf{default} & \textbf{description} \\ 
 \hline
 $\sigma$ & $\sfrac{103}{100}$ & position of the decay near pseudo-median \\ 
 \hline
 $\lambda$ & $\sfrac{20}{100}$ & width of the decay (in \% of side lengths) \\ 
 \hline
 $\delta_1$ & $\sfrac{20}{100}$ & width of transition between distances on left and right sides \\ 
 \hline
 $\delta_2$ & $\sfrac{20}{100}$ & width of transition between varying widths on left and right sides \\ 
 \hline
 $\delta_3$ & $\sfrac{20}{100}$ & width of transition between edge distance and median distance \\ 
 \hline
 $\sigma_3$ & $\sfrac{20}{100}$ & position of transition between edge distance and median distance \\ 
 \hline
\end{tabular}
\caption{List of control parameters for the weight functions.}
\label{Table_weight_functions}
\end{table}

\begin{figure}[ht!]
  \centering
        \subfigure[ ]{\includegraphics[width=0.49\linewidth]{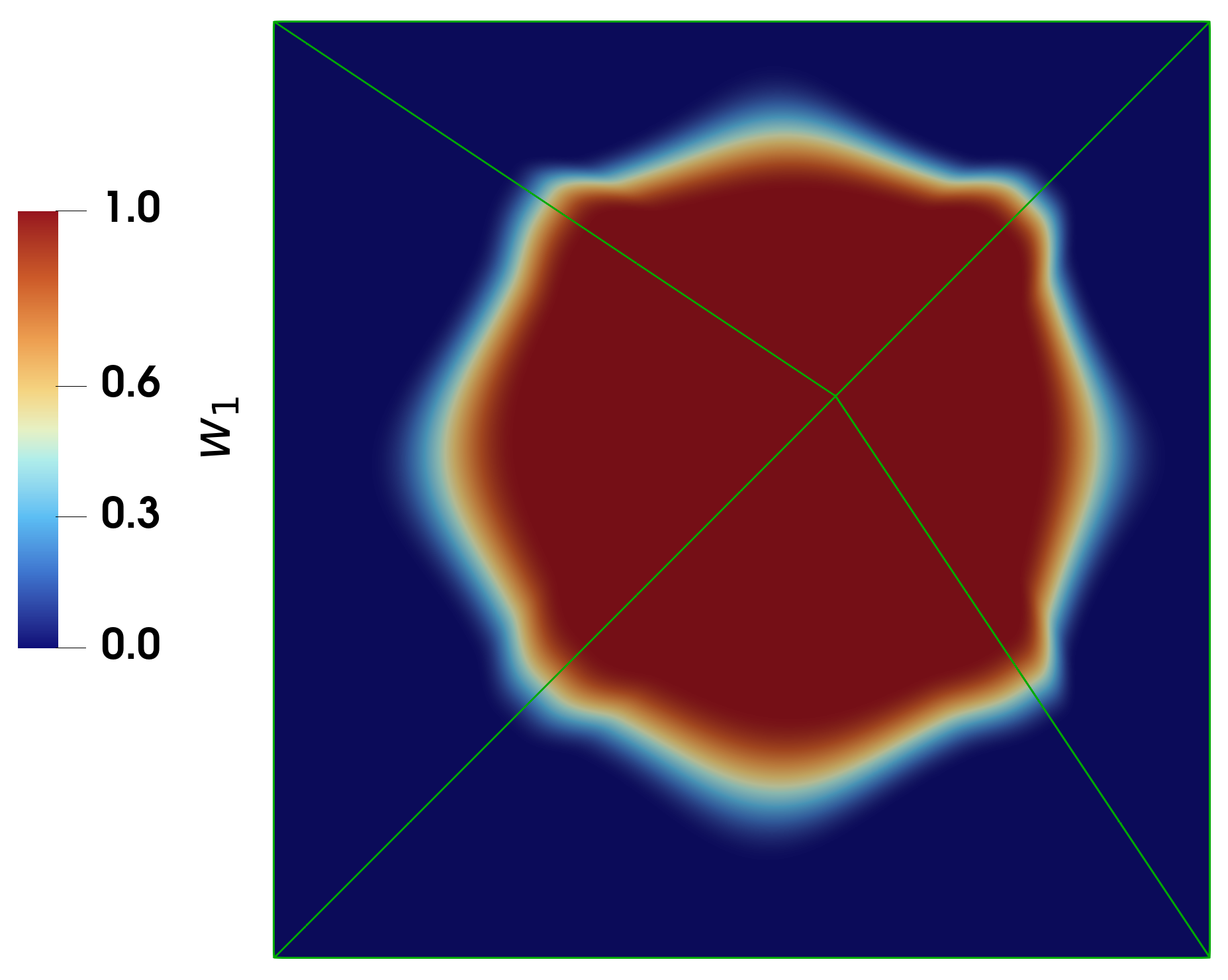}}
        \subfigure[ ]{\includegraphics[width=0.49\linewidth]{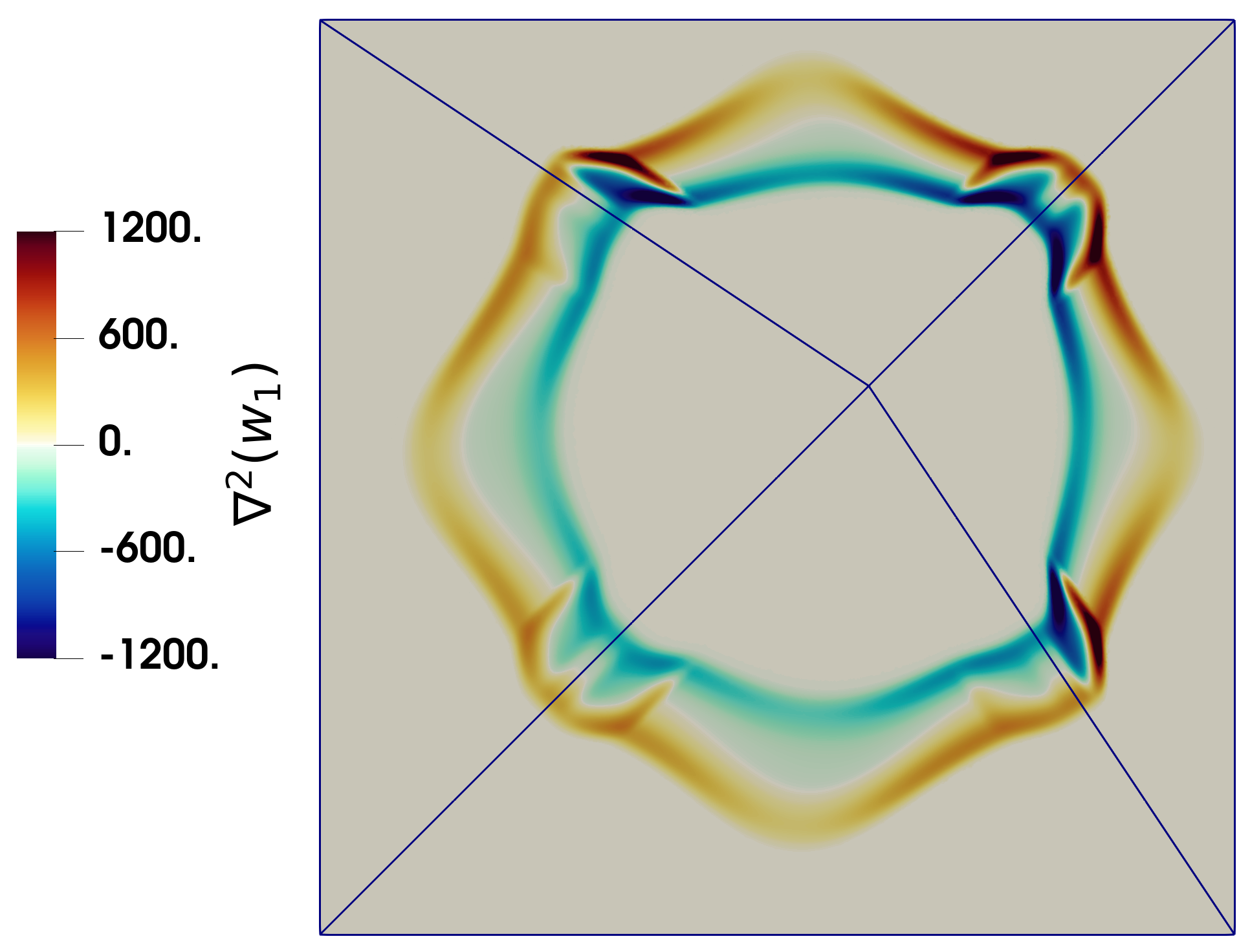}}
  \caption{Ring of 4 elements around a central node, showing \textnormal{(a)} the weight of the central node, $w_1(x,y)$, \textnormal{(b)} the Laplacian $\nabla^2(w_1)$, demonstrating smoothness of the weight function and its derivatives across element edges.}
  \label{Figure_weight_derivatives}
\end{figure}

Figure-\ref{Figure_weight_functions} shows a single triangular element with the distribution of the first weight function $w_1$, as well as the sum of all three weight functions before they are normalised, to illustrate the overlap regions. Towards the edges of the element, the overlap region receives contribution from the two adjacent nodes, while in the centre of the element, all three nodes are contributing to the full solution equation-(\ref{nodal_formulation2}). Likewise, Figure-\ref{Figure_weight_derivatives} shows the first weight function $w_1$ on a ring of 4 elements around a central node, illustrating how the function smoothly transitions across element edges. To provide a qualitative measure of the smoothness across edges, the Laplacian $\nabla^2(w_1)$ is also plotted to show the continuity of second-order derivatives.

\subsubsection{FEM-NN integration and training loss}

In order to train all the NN's of a given FEM-NN grid, the solution must be integrated around all parent elements of a given node. In PINNs, the usual integration method is typically the MSE, whereas in classical FEM methods this might be a weak-form spatial integration. For equation-\ref{Crank-Nicolson}, at a given time $t_n$, the residual map on element $j$ can be expressed as an operator acting on the Neural Network $\mathcal{N}_{i,k,n+1}$ being trained,
\ba
  \mathcal{R}_j\Big|_{i,k}(\mathcal{N}_{i,k,n+1}) &=& \mathbf{u}_{j,n+1}\Big|_{i,k}(\mathcal{N}_{i,k,n+1}) \ttt{ }-\ttt{ } \mathbf{u}_{j,n} \nonumber\\
      &&-\ttt{ } \frac{\Delta t}{2} \left[F(\mathbf{u}_{j,n+1}\Big|_{i,k}(\mathcal{N}_{i,k,n+1})) + F(\mathbf{u}_{j,n})\right], \label{Residual_map}
\ea
where the function $\mathbf{u}_{j,n+1}\Big|_{i,k}(\alpha)$ is obtained by substituting the scalar map $\alpha$ into the relevant component of the multi-variate FEM-NN solution, as formulated by equation-(\ref{nodal_formulation2}), such that
\be
  \mathbf{u}_{j,n+1}\Big|_{i,k}(\alpha)
  =
  \begin{pmatrix}
    u_{j,1,n+1} \\
    u_{j,2,n+1} \\
    ... \\
    u_{j,k,n+1}\Big|_i(\alpha) \\
    ... \\
    u_{j,m,n+1}
  \end{pmatrix}
  =
  \begin{pmatrix}
    \sum_l w_l\mathcal{N}_{l,1,n+1} \\
    \sum_l w_l\mathcal{N}_{l,2,n+1} \\
    ... \\
    w_i \alpha + \sum_{l\ne i} w_l\mathcal{N}_{l,k,n+1} \\
    ... \\
    \sum_l w_l\mathcal{N}_{l,m,n+1} \\
  \end{pmatrix}
\ee
This residual map equation-(\ref{Residual_map}) now needs to be integrated on all the parent elements of the node $i$. The integral operator is thus defined as 
\be
  <\alpha,\beta>_{i,k} \ttt{ }=\ttt{ } \frac{1}{n_p(i)}\sum_{j=1}^{n_p(i)} \int_{\mathcal{D}_j} \mathcal{R}_j\Big|_{i,k}(\alpha) \ttt{ } \mathcal{R}_j\Big|_{i,k}(\beta) \ttt{ } d\mathbf{X} \label{Integral_operator}
\ee
where $\mathcal{D}_j$ is the spatial domain of a given element $j$, and where the summation $1\le j\le n_p(i)$ is done over the $n_p$ parent elements of the $i^{th}$ node. Essentially, the integral $\int_{\mathcal{D}_j}(.)d\mathbf{X}$ on each element is calculated as the MSE over a random sampling of points on the element $j$. This provides the final formulation of training loss for each PDE variable/equation $k$ at a given node $i$, as
\be
  \mathcal{L}_{i,k} \ttt{ }\ttt{ }=\ttt{ }\ttt{ } <\mathcal{N}_{i,k,n+1},\ttt{ } \mathcal{N}_{i,k,n+1}>_{i,k},\label{Loss_function}
\ee

The reason for this specific choice of formulation, expressing the integral loss as a binary operator on two identical operands $\mathcal{N}_{i,k,n+1}$, will become evident when describing the ENGD optimisation algorithm in Subsection-\ref{Subsection_ENGD}.

Although equation-\ref{Residual_map} is more intuitive to perceive in this time-increment form, with $\mathbf{u_{n+1}} - \mathbf{u_{n}}$, it is actually not exact, or at least not fully generic. That's because the time-derivative term of an equation may itself be a more complex function of multiple variables, such as $\rho\partial_t(\vec{v})$ for a momentum equation, or $\partial_t(\rho T)$ for an energy equation. Likewise, some equations in a PDE system may be time-independent, for example when using an auxiliary variable with a Poisson equation. Therefore, the fully generic formulation of the residual map equation-(\ref{Residual_map}) at FEM-NN node $i$ and element $j$ for variable $k$ can be expressed as
\ba
  \mathcal{R}_j\Big|_{i,k}(\mathcal{N}_{i,k,n+1}) &=& f\left(\mathbf{u}_{j,n}, \ttt{ } \mathbf{u}_{j,n+1}\Big|_{i,k}(\mathcal{N}_{i,k,n+1}) \right)\nonumber\\
      &&-\ttt{ } \frac{\Delta t}{2} \left[F(\mathbf{u}_{j,n+1}\Big|_{i,k}(\mathcal{N}_{i,k,n+1})) + F(\mathbf{u}_{j,n})\right], \label{Residual_map2}
\ea
where $f$ represents the time-derivative term of the PDE equations.

\subsubsection{Boundary Conditions}

If an element $j$ is located at a boundary of the domain, then a boundary loss term needs to be added to the training. The full formula for the training loss then becomes
\be
  \mathcal{L}_{i,k} \ttt{ }\ttt{ }=\ttt{ }\ttt{ } <\mathcal{N}_{i,k,n+1},\ttt{ } \mathcal{N}_{i,k,n+1}>_{i,k} \ttt{ }+\ttt{ }\ttt{ } \zeta <\mathcal{N}_{i,k,n+1},\ttt{ } \mathcal{N}_{i,k,n+1}>^{b}_{i,k},\label{Loss_function2}
\ee
where the boundary integration is applied to the boundary operator as
\ba
  <\alpha,\beta>^{b}_{i,k} &=& \frac{1}{n_p(i)}\sum_{j=1}^{n_p(i)} \int_{\mathcal{B}_j} S_j\Big|_{i,k}(\alpha) \ttt{ } S_j\Big|_{i,k}(\beta) \ttt{ } d\mathbf{X} \label{Boundary_integral_operator}\\
  S_j\Big|_{i,k}(\alpha) &=& G\left(\mathbf{u}_{j,n+1}\Big|_{i,k}(\alpha)\right) - g.
\ea
with $\mathcal{B}_j$ being the boundary edges of the element $j$.

\subsubsection{Central versus Global representation of variables}

A fundamental aspect of definition-(\ref{nodal_formulation}) is the fact that, just like barycentric coordinates, the weight functions $(w_1,w_2,w_3)$ are also normalised so that they sum up to one everywhere on their element. As depicted in Figure-\ref{Figure_weight_derivatives}, the solution of a given node's NN is designed to be valid on all its parent elements. Likewise, this NN's validity may also extend beyond its delimited weight region (the red region in Figure-\ref{Figure_weight_derivatives}a). Thus, if each node's NN is capable of representing, on its own, a scalar field $u_{j,k,n}$ on the entire parent element, then it follows that
\be
   \mathcal{N}_{1,k,n} = \mathcal{N}_{2,k,n} = \mathcal{N}_{3,k,n} = u_{j,k,n} = w_1\mathcal{N}_{1,k,n} + w_2\mathcal{N}_{2,k,n} + w_3\mathcal{N}_{3,k,n}.
\ee
This property of the Fuse-PINN formulation is a pivotal characteristic of the code's numerical stability, because it allows a training formulation which ensures the solution $u_{j,k,n}$ is evenly shared between the three nodes of an element. Whereas equation-(\ref{nodal_formulation}) represents the \textit{global} solution on a given element with contributions from all three nodes, consider the \textit{central} solution on an element (or a ring of parent elements around a central node) given by the central node's NN, without any weight function, as
\be
  \hat{u}_{i,k,n} = \mathcal{N}_{i,k,n}.\label{nodal_formulation_central}
\ee
Note the distinction between the indexing of $u_{j,k,n}$ and $\hat{u}_{i,k,n}$, which is intentional: one variable is evaluated on an element $j$ (with contributions from its three nodes), while the other is evaluated around a single node $i$ on any parent element. This alternative solution representation can be used to impose an evenly distributed solution on the elements, by reformulating equation-(\ref{Residual_map2}) as
\ba
  \mathcal{R}_j\Big|_{i,k}(\mathcal{N}_{i,k,n+1}) &=& f\left(\mathbf{u}_{j,n}, \ttt{ } \hat{\mathbf{u}}_{i,n+1}\Big|_{k}(\mathcal{N}_{i,k,n+1}) \right)\nonumber\\
      &&-\ttt{ } \frac{\Delta t}{2} \left[F(\mathbf{u}_{j,n+1}\Big|_{i,k}(\mathcal{N}_{i,k,n+1})) + F(\mathbf{u}_{j,n})\right], \label{Residual_map3}
\ea
ensuring that the solution at the next time-step is both evenly distributed and coherent between all nodes, while retaining information from all neighbouring nodes imposed through all other terms in the PDE equations.

This \textit{central equaliser} formulation is particularly essential to avoid one node NN taking over all other nodes on an element. Indeed, when simulating models with rapidly convecting structures, we observed that without this numerical trick the first node to receive the convected structure would ``monopolise'' the solution on its element, forcing the other two node's NN contributions to zero, and compensating for the decaying weight by augmenting the NN value itself. This works fine until the convection reaches the edge of the overlap region: the other two nodes having been suppressed to zero, they hold no information about the oncoming convection, resulting in a sort of boundary effect preventing the convection from travelling further. Mixing the central and global formulation of variables by introducing the central formulation in the time-derivative resolves this issue, allowing solutions to always be fairly distributed between all nodes of a given element.

To the best of our knowledge, this concept of \textit{global} versus \textit{central} FEM-NN formulation, with this \textit{central equaliser} trick, is characteristic of the nature of our Fuse-PINN proposition, since clearly no equivalent can be achieved with classical FEM formulations. The surprising simplicity of this \textit{central equaliser} as presented in equation-(\ref{Residual_map3}), and the simplicity of its implementation, betrays the lengthy and difficult exploration that was required to reach this novel, non-trivial conclusion.

The only exception when this \textit{central equaliser} is not used is for equations that are time-independent and implicit in nature. This is the case of the Poisson equation, which is often used for auxiliary variables, typically for vorticity terms, with $\nabla^2\phi = \omega$, solving for the variable $\phi$. Here, the PDE form of equation-(\ref{Residual_map3}) would have $f_k = \omega_{n+1} - \nabla^2\phi_{n+1}$ and $F_k = 0$. However, for implicit equations like these, it is essential to retain the full \textit{global} FEM-NN formulation of both $\phi$ and $\omega$, otherwise information from the boundary conditions at the edge of the domain cannot be communicated across elements.

\subsubsection{GMSH grid generation}

The advantage of a fully generic triangular FEM-NN approach is that it allows coupling with other meshing libraries like GMSH \cite{GMSH,GMSHweb}. Although the coupling is currently still minimal with simply 2D mesh generation inside a polygon, this proof-of-concept can easily be extended to more complex use cases.

\begin{figure}[ht!]
  \centering
        \subfigure[ ]{\includegraphics[width=0.49\linewidth]{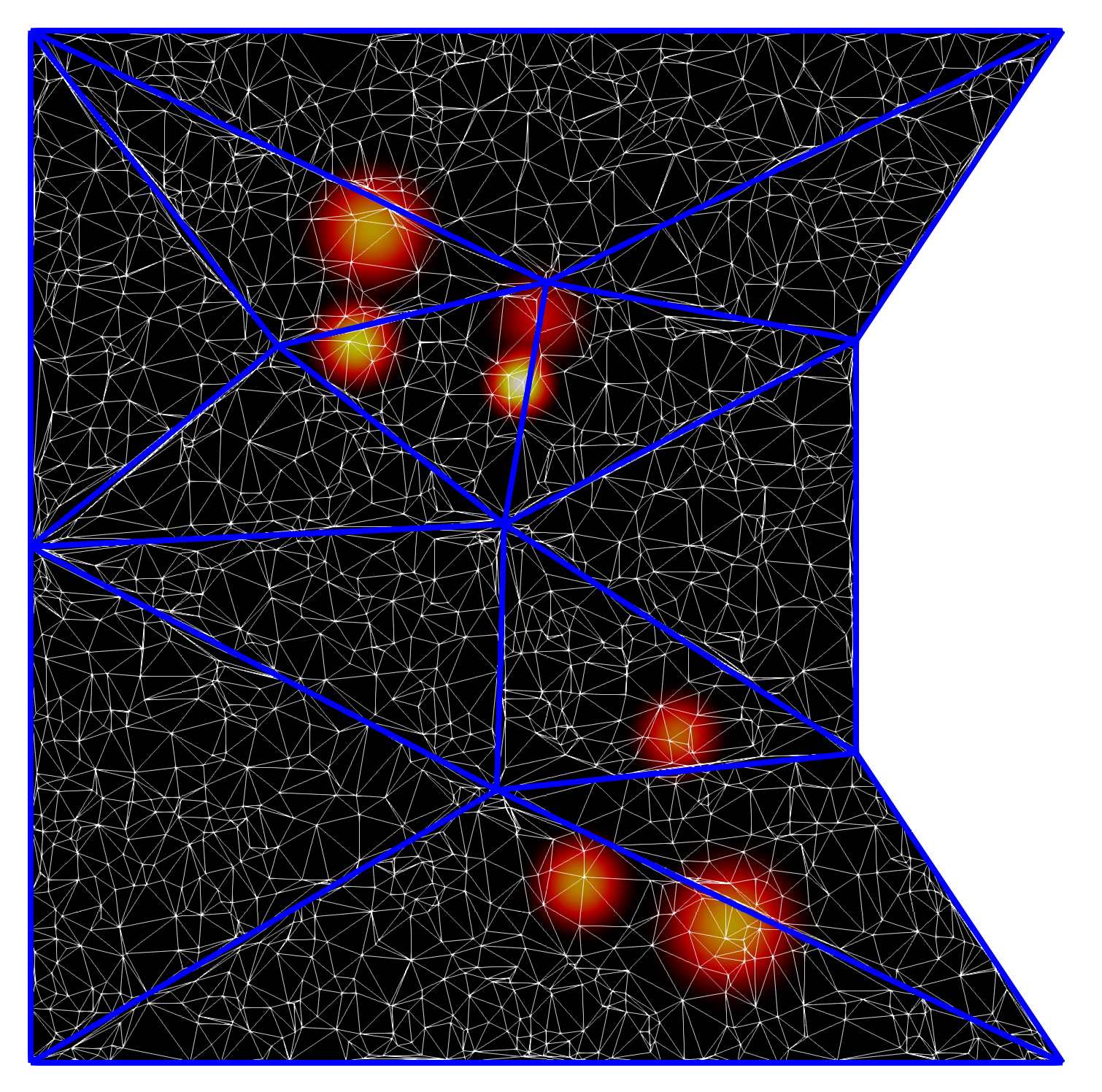}}
        \subfigure[ ]{\includegraphics[width=0.49\linewidth]{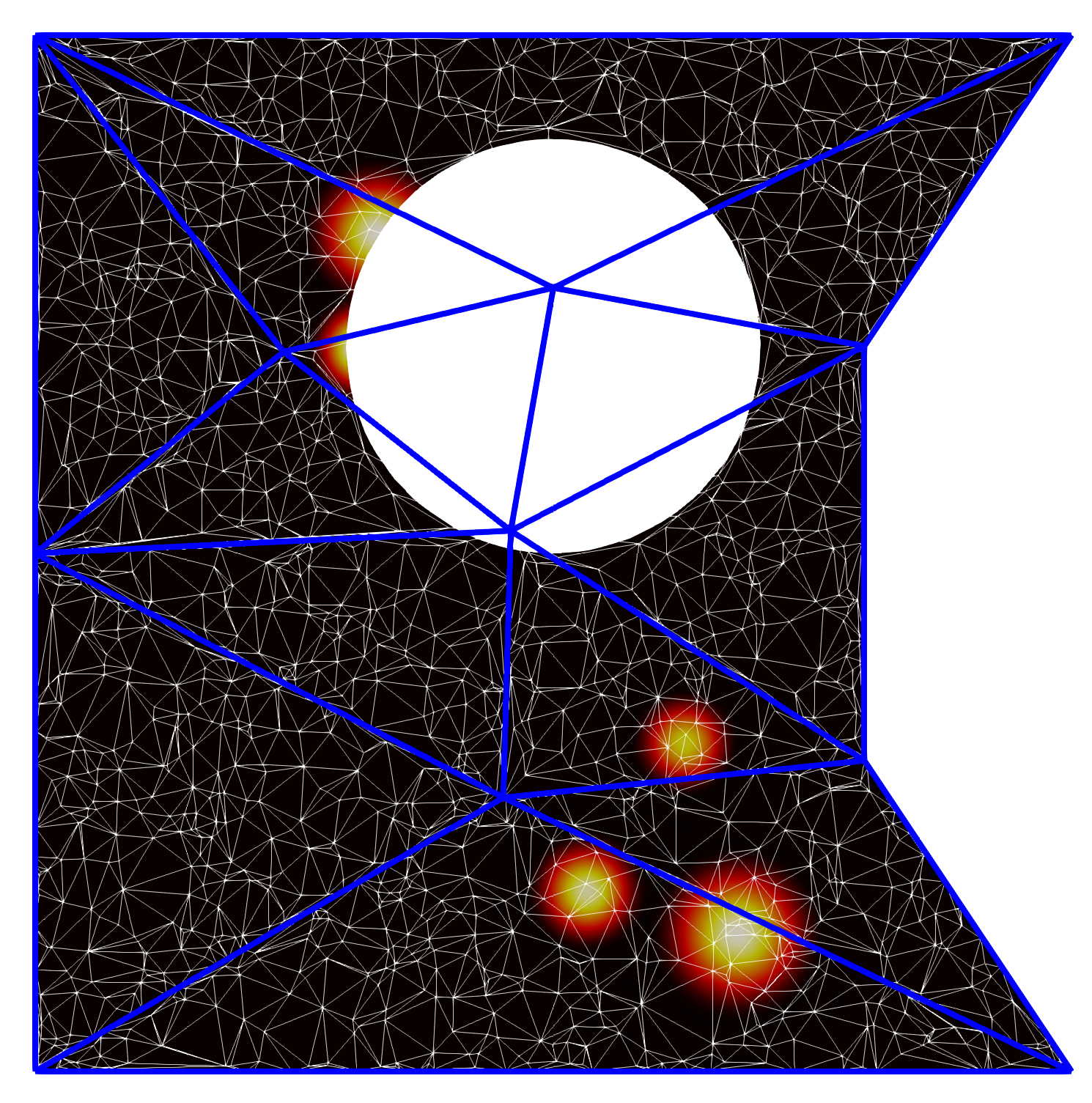}}
  \caption{Examples of grids generated with GMSH \textnormal{(a)} with a user-defined boundary contour, \textnormal{(b)} with an arbitrary obstacle in the middle of the grid. The grids contains 13 elements with 11 nodes in total, hence 11 PINNs for each physics variable. The background color shows an arbitrary variable being solved for, shown here for context. The blue lines are the edges of the elements. The white lines show the random sampling used inside each element for the PINN training. Higher-resolution sampling is typically used for simulations, but low-resolution was used here for illustration purposes. Note that obstacles with boundary conditions can be placed anywhere on the grid, including above grid nodes.}
  \label{FIGURE_gmsh}
\end{figure}

Figure-\ref{FIGURE_gmsh}a shows an example of a grid generated using GMSH inside a user-prescribed contour. The user chooses a number of nodes they would like to use, and the GMSH grid generator is run with varying length-scales until a grid is found with the right number of FEM-NN nodes. The grid shown in this figure contains 13 elements with 11 nodes in total. This means 11 Neural Networks are being trained for each physics variable in the PDE system. The background color shows an arbitrary variable being solved for, shown here for context. The blue lines show the edges of the elements, while the white lines show the random sampling used inside each element for the PINN training. Higher-resolution sampling is typically used for simulations, but low-resolution was used here for illustration purposes.

\subsubsection{Mesh-free compromise}

One of the great advantages of PINNs that's often brought forward is their mesh-free aspect. Therefore, imposing a mesh structure in the underpinning design of the Fuse-PINN framework may be considered as a step backwards. However, this approach is a pragmatic step towards a robust parallelisation method which, to the best of our knowledge, has never been proposed before. A key aspect of the Fuse-PINN implementation is that it still allows arbitrary boundary conditions across several elements, allowing for obstacles or foils or other structures to be inserted into the calculation, without affecting the FEM-NN approach itself. In other words, the FEM-NN approach allows for parallelisation, without restricting arbitrary meshing structures.

The implementation of obstacles inside the code is in fact extremely simple, it only requires additional boundary points around the obstacles themselves, while domain points that are inside the obstacle are masked out. Figure-\ref{FIGURE_gmsh}b shows an example of such a grid with a simple circular obstacle. Note that obstacles can be placed anywhere on the grid, including above grid nodes. The PINNs associated to the obstructed node will still contribute to the elements around it that have sampled points outside the obstacle.

\subsection{Normalisations of the PINN's inputs/outputs and residuals}

The spatial grid (i.e. the PINN's input) is normalised to $[-1,+1]$, while physics variables (i.e. the PINN's outputs) are simply normalised with respect to a user-defined constant (i.e. just a rescaling, no shift). Other options are available, either to normalise variables with the maximum amplitude of the initial conditions across the entire domain, or with the amplitude of the solution at the previous time-step, locally on each element.

The outputs of the PINNs on each node are of course de-normalised before they enter the PDE residuals for each equation. Additional normalisation is then used for these PDE residuals. In most cases, this normalisation is straight-forward and corresponds to the variable being solved in the equation. For example, for the continuity equation, which solves for $\partial_t\rho$, the equation residual is normalised with the density's normalisation factor. For other equations, more complex normalisations may be appropriate, for example with the momentum equation, if a pressure term $\nabla p$ is the driving term, then it is best to normalise with respect to the pressure.

Note that the normalisation of the PDE residuals isn't just meant for diagnostic purposes to get more human-readable losses. The choice of normalisation can have an important effect on the convergence of the training, particularly with the Energy Natural Gradient Descent optimiser. Certainly, it is essential to ensure that normalisation is consistent in all components of the preconditioner when computing the Energy Natural Gradient Descent.

\subsection{Energy Natural Gradient Descent (ENGD) Optimiser}\label{Subsection_ENGD}

The remarkable work by \cite{Muller2023} is a core component of Fuse-PINN, without which our results would almost certainly not be presentable. While typical optimisers like Adam and L-BFGS are also available in Fuse-PINN, in practice they are simply not accurate enough nor fast enough to be useful for a production tool. In contrast, the ENGD implementation allows for faster, higher-accuracy solutions and, consequently, long time-integration.

The trade-off of ENGD is that it is a physics-based optimiser, meaning it cannot be implemented as a fully modular library in frameworks like PyTorch or Jax. However, as will be described below, our implementation was made as modular as possible, such that users can focus only on the PDE equations, without having to worry about the ENGD implementation at all.

\subsubsection{ENGD summary}

The ENGD optimiser can be summarised as applying a physics preconditioner to the model gradients. It requires building and inverting the so-called Gram matrix, which is described in details in \cite{Muller2023}, and summarised quickly here for convenience. In its simplest description, the ENGD optimiser can be described with the Algorithm-\ref{ENGD_algorithm}.
\begin{algorithm}
  \caption{ENGD optimiser}\label{alg:cap}
  \scriptsize
  \begin{algorithmic}[1]
    \REQUIRE $X$           \COMMENT{The spatial domain}
    \REQUIRE $\mathcal{N}$ \COMMENT{The Neural Network for the physics variable $u$}
    \REQUIRE $\mathcal{R}$ \COMMENT{The PDE residual map as an operator on $u$}
    \FOR{$iter \le n_{epoch}$}
      \STATE $\mathcal{N} \gets$                                       \COMMENT{Retrieve NN model from previous epoch}
      \STATE $u \gets X, \mathcal{N}$                                  \COMMENT{Evaluate physics variable from NN}\label{ENGD_var1}
      \STATE $\partial u \gets X, u$                                   \COMMENT{Evaluate all spatial derivatives of variable}\label{ENGD_var2}
      \STATE $L(\theta) \gets \int \mathcal{R}(u)^2$                   \COMMENT{Integrate residual over domain to obtain loss}\label{ENGD_loss1}
      \STATE $\nabla L(\theta) \gets L$                                \COMMENT{Backward step to get model gradients}
      \STATE $\tilde{u}_i \gets \partial_{\theta_i} u$                 \COMMENT{Variable derivatives w.r.t. each model parameter}\label{ENGD_deriv1}
      \STATE $\partial\tilde{u}_i \gets \partial_{\theta_i}\partial u$ \COMMENT{Same for all spatial derivatives of $u$}\label{ENGD_deriv2}
      \STATE $G_{i,j} \gets \int \mathcal{R}(\tilde{u}_i)\mathcal{R}(\tilde{u}_j)$ \COMMENT{Assemble Gram matrix}\label{ENGD_Gram}
      \STATE $\nabla^E L(\theta) \gets G^{-1}_{i,j}\nabla L(\theta)$ \COMMENT{Correct gradients with Gram inverse}
      \STATE $\eta^{\ast} \gets arg min_{\eta\in[0,1]} L[\theta - \eta\nabla^E L(\theta)]$ \COMMENT{Line search for best learning rate}\label{ENGD_linesearch}
      \STATE $\theta \gets \theta - \eta^{\ast}\nabla^E L(\theta)$ \COMMENT{Update model parameters}
    \ENDFOR
  \end{algorithmic}\label{ENGD_algorithm}
\end{algorithm}

This algorithm is thus applied, in parallel, for each variable $k$, for each node $i$, integrating over the node's parent elements $j$, such that the algorithm's inputs are taken to be
\ba
  X &\in& D_j \\
  \mathcal{N} &=& \mathcal{N}_{i,k,n+1}\\
  \mathcal{R} &=& \mathcal{R}_j\Big|_{i,k}
\ea

\subsubsection{Implementation of the Gram matrix assembly}

The Gram matrix construction is by far the heaviest part of the ENGD computation, in terms of both compute and memory. In particular, steps \ref{ENGD_deriv1} and \ref{ENGD_deriv2} from Algorithm-\ref{ENGD_algorithm} require auto-differentiation of the Neural Network $\mathcal{N}_{i,k,n+1}$ with respect to each model parameter, as well as all spatial coordinates (in our case $x$ and $y$) up to second order -- at least for most physics models with diffusive Laplacian terms. This construction of the NN derivatives is therefore vmap'ed in PyTorch to ensure high efficiency.

However, this step results in a substantial graph which can set a hard limit on resolution. For cases with a large number of sample points per element, combined with a large NN model, the Gram construction can lead to issues with memory overload. To avoid this limitation, the integration of each element sample points can be split into batches. Indeed, since all sample point contributions are summed up, as long as the total number of points is tracked, this can be batched and looped over sequentially. Obviously, the counterpart is longer compute time. A user-input parameter is available to control the number of batches for splitting the ENGD integration, with a default of 1. One of the key hyper-parameter tuning of FEM-NN is to find the right balance between the total resolution, the number of grid nodes, and the size of the NN model, to stay below this memory limit.

\subsubsection{Gram matrix regularisation}

The Gram matrix is generally not invertible directly. Alternative methods, such as pseudo-inverse or least-square solution, are more reliable. Even then, the Gram matrix may include zero diagonal elements as well as degenerate entries (NaNs). These are regularised with a user-input parameter, typically ranging between $10^{-14}$ and $10^{-6}$.

A higher value for the Gram matrix regularisation will lead to a faster initial drop in the loss -- it helps the algorithm find the minimum well in the loss landscape quickly. However, if the regularisation is too high, it may also prevent the algorithm from achieving high accuracy. At present, we have not implemented a scheduled decay of the regularisation, because in practice, if the regularisation parameter is small, then the first time-step may take slightly longer to converge, but subsequent time-steps are fast anyway, because the previous time-step NN's are already very close to the solution (assuming the solution change between two time-step is reasonable).

\subsubsection{Line-search implementation}

The line-search for the best learning rate $\eta^{\ast}$ could potentially be parallelised, but its cost is negligible compared to the Gram matrix construction. Instead of a blunt grid search, an incremental search is implemented to reduce the number of loss evaluations, with successive grid searches (or Brent's method), progressively refining the range of $\eta$ to be tested, thus reducing the total number of evaluations while narrowing down to a precise final value for $\eta^{\ast}$.

User-input parameters are available to control the parameters of the line-search, namely
\ba
  \eta_{min} &:& \textnormal{minimum value of the line-search range}, \nonumber\\
  n_{cycles} &:& \textnormal{number of successive cycles of narrowing intervals}, \nonumber\\
  n_{eval}   &:& \textnormal{number of evaluations at each cycle}. \nonumber
\ea

\subsubsection{Equation linearisation}

A key aspect of the ENGD optimisation algorithm is the requirement for the binary operator formulation of equation-(\ref{Loss_function}) to be applied to different functions: to the Neural Networks $\mathcal{N}_{i,k,n+1}$ for steps \ref{ENGD_loss1} and \ref{ENGD_linesearch} of Algorithm-\ref{ENGD_algorithm}, and also applied to the model derivatives $\tilde{u}_i$ in step \ref{ENGD_Gram}. Effectively, this means that the physics equations need to be linearised with respect to the variable being trained.

While this may seem complex to implement, the current choice of Fuse-PINN to use separate NN's for each physics variable means that this linearisation only needs to be done for a single variable per equation. While a combined approach, with a single NN for all variables, may offer a more compact implementation of the PINN, its disadvantage would be that each equation would need to be linearised with respect to every variable, significantly increasing the complexity of the physics model implementation.

\subsubsection{Variable caching}

Since the loss function equation-(\ref{Residual_map3}) at node $i$ and variable $k$ involves other non-trainable variables, including the $k$ variable NN's from neighbour nodes, there is no need to recalculate these repeatedly for the Gram construction and the line-search. Therefore, these variables are cached (and detached) at the first loss calculation in steps \ref{ENGD_var1},\ref{ENGD_var2},\ref{ENGD_loss1} of Algorithm-\ref{ENGD_algorithm}, which reduces the compute time of each epoch.

\subsection{Parallelisation and Communications}

The main purpose of Fuse-PINN is to demonstrate the viability of PINNs for real-world application, at scale, on High-Performance-Computing (HPC) infrastructures. It leverages this novel FEM-NN formulation, combined with other numerical methods like implicit time-schemes, Energy Natural Gradient Descent, and of course MPI communications.

In this implementation, a multi-level parallelisation of the workflow allows for scaling out to high resolutions at minimal cost, while almost all inter-GPU communications remain local to each FEM-NN node, regardless of resolution, ensuring performance scaling. There are two levels of parallelism: on variables and on FEM-NN nodes. Since each node has separate Neural Networks for each physics variable, it is natural to allocate one GPU per Neural Network.

The user may also choose to run sequentially with all Neural Networks on a single GPU, or with the parallelisation applied only to variables. It is also possible to run with a single Neural Network for the entire spatial domain (ie. without using FEM-NN).

Since the integration of the PDE loss function for a given equation $k$ at node $i$ requires more than just the variable being trained (in some equations, all variables are involved), this means each GPU requires a copy of the other variable NN's, not only from the given node $i$, but from all neighbour nodes as well, in order to reconstruct the full solution on all parent elements. In addition, each GPU also requires a copy of these NN's from the previous time-step.

At the end of each epoch, when the NN's on all GPUs are updated with new parameters $\theta$, these new models need to be shared between neighbour nodes, for each variable NN GPU. This enables the full solution at the next epoch to be obtained with the latest NN models on all parent elements around each node $i$. This is done by gathering, at node $i$, all the relevant NN models from neighbour nodes, for all variables. Thus, regardless of the size of the FEM-NN grid, communications at each epoch of a given time-step are always local to each node $i$. Indeed, a given node does not require the NN from nodes that are beyond its direct neighbours. However, information still travels across the grid owing to the FEM-NN formulation with the overlap region on each element. This aspect is critical for implicit solves like Poisson problems, which can still be achieved with calculations local to each node.

There are only two global communications involved in the implementation. The first one is to gather the final loss value from all nodes for each variable. These individual node losses are gathered on the master GPU and averaged across the whole grid, so that it can be logged and printed. The second one is when the full solution needs to be either plotted, or saved to disc for checkpointing. In that case, all NN's from all GPUs are gathered on the master node which post-processes these accordingly. Separate input parameters allow the user to obtain these loss-logs, plots and checkpoints either at fixed epoch intervals, or simply at the end of a time-step. In cases where the grid contains a large number of FEM-NN nodes with large NN models, gathering all the NN's on the master node and plotting the solution can be time-consuming, and is best kept for the end of the time-step.

\subsection{User-inputs and Post-Processing}

The Fuse-PINN user can provide input parameters directly as a Python input file, or as command-line arguments. If a user-input parameter is provided in the input file, it can be overwritten in the command-line directly to allow easier exploration of parameters. This allows users to rely on a baseline input file with all the required inputs, while exploring other parameters more freely in the command-line.

The usual Weights \& Biases logging hooks are not implemented in Fuse-PINN. W\&B requires a web-service running on the computing architecture, which is typically unavailable on most research HPC systems (for security reasons). Instead, all losses for each variable on the domain and the boundary conditions are logged on file, which can be used to plot the progress of a given time-step at any point, for which a user script is available. The losses can be recorded at any epoch interval the user chooses. Likewise, they can be printed out to the terminal output at any chosen interval.

Plots of the variables can also be created at chosen epoch intervals to follow the progression of a time-step convergence. These are simple Matplotlib figures saved to disk. For more advanced plots, the solution can be extracted to VTK format, allowing visualisation in Paraview. For all plots, the default sampling resolution on each element is set to the same resolution as the one used for the residual loss integration, but it can be lowered to allow for smaller plot files.

\subsection{Physics Models}

As explained earlier, the implementation of the various physics models is kept as modular as possible, to allow for a user-friendly development framework. The user does not need to worry about the ENGD optimiser or the details of the epoch loops or checkpointing etc., it is entirely isolated. The Initial and Boundary Conditions functions are part of the physics class, as these may vary significantly between physics models. To give an idea of the coding involved for a new physics model, a PDE system with four variables involves only about 500 lines of code, including initial and boundary conditions.

As may be seen from the Crank-Nicolson formulation in equation-(\ref{Crank-Nicolson}), the RHS of the PDE, namely the function $F$, needs to be calculated for both the previous and current time-step variables. This requires a dedicated function for the RHS calculation, which is called by the master PDE function.

Also, given the requirement of the ENGD optimiser for a linearised equation, the user needs to write the full PDE equations and their linearised version with respect to the variable being trained for its corresponding equation. However, since this is only needed for a single variable per equation, it remains relatively lightweight.

Each physics model is described in more details in the relevant results section below.

\subsection{Limitations}

\subsubsection{Soft BC's}

One of the first limitations of the current Fuse-PINN implementation is that it only includes ``soft'' boundary conditions, such that a residual loss term for the boundary conditions is added to the main PDE domain loss (with a user-input scaling factor). While this works well, other PINN studies have demonstrated successful implementations of ``hard'' boundary conditions, where the boundary values of the NN solution is enforced through the architecture of the solution itself, for example with a tanh at the edge which forces the solution to be zero on the boundary.

While this should be relatively straight-forward to implement, it was not considered essential for the demonstrative purpose of the current work. However, it may be highly preferable for more complex physics models with more challenging boundary domains and boundary equations, and should be considered for future improvements of the Fuse-PINN framework.

\subsubsection{Positivity of variables}

Another advantage of PINNs is that the architecture of the Neural Network can be designed to enforce positivity of (some) variables. For example, if it is sure that density and temperature should remain positive everywhere, this may help avoid negative values which can trigger numerical instabilities inside certain physics models. This positivity option is implemented in Fuse-PINN, but unfortunately it was found to strongly affect the performance of the solver convergence. Even for cases where variables remained positive anyway, the performance was strongly degraded. Further investigation will be needed to explore more elaborate positivity NN architectures to understand whether this can be achieved without degrading convergence.

\subsubsection{Overkill for small simulations}

Counter-intuitively, our FEM-NN approach is not competitive for small use cases, where traditional FEM solvers remain more effective. However, when it comes to large spatial domains with high resolutions, it becomes interesting. Detailed scaling tests are presented in the Section-\ref{Section_scalings} to identify quantitatively how well the FEM-NN architecture scales towards challenging, high-resolution use-cases, and how it compares to classical numerical solvers.

\subsubsection{Curved edges, 3D grid, hp-refinement, and localised sampling}

Further developments of the grid would be desirable to enable more elaborate spatial domains. As shown in Figure-\ref{FIGURE_gmsh}b, arbitrary obstacles (and therefore boundaries) are easy to include, which may be sufficient to cover all exotic boundary designs. However, there may be advantages to allow curved triangular element edges, which would be relatively simple to implement, as it only requires a re-mapping of the local barycentric coordinates. For example, this might be numerically more precise when combined with ``hard'' boundary conditions.

Implementation of 3D grids remains to be addressed, but the FEM-NN formulation would be identical. The only challenging part would be to generalise the weight functions described in Section-\ref{Section-Weight-Functions}. Still, this is a significant piece of work which would be required to advance Fuse-PINN towards realistic 3D applications.

Similar to Finite-Element Methods, the FEM-NN implementation in Fuse-PINN could easily be generalised to allow for $h^p$ refinement. In the case of FEM-NN, $h$-refinement is identical to classical FEMs, where the number of grid nodes is increased/decreased to adapt the spatial resolution to specific areas of the domain, and this is of course already possible in the current implementation. The $p$-refinement, however, would require changing the size of the NN at given grid nodes, potentially even with varying NN sizes for different physics variables at a given node. This would allow higher NN resolution at specific domain areas where the solution might be more challenging. Nevertheless, an important aspect of such $h^p$ refinement is that it would be best implemented together with a more elaborate parallelisation of the grid nodes. If all grid nodes (and variables) are allocated their own GPU, then the entire calculation will be as slow as the largest NN on the grid, with other smaller NN grid points waiting for that solution to come through. Therefore, $h^p$ refinement would only make sense if multiple small-NN grid nodes can be grouped together on a single GPU, while other large-NN grid nodes have their own GPU allocated. This feature is not available at present and would require significant re-implementation of the parallelisation and communications.

Localised sampling would also be much more efficient. At present sampling is done uniformly across all elements. Increasing the sampling density where the simulation dynamics is strongest would allow for cheaper compute costs overall. However, this would only be coherent if combined with localised FEM-NN node location, otherwise the time-to-solution would be bounded by the FEM-NN element with the most sampling points. The integration of Fuse-PINN with GMSH provides a direct route to localised FEM-NN node location.


\section{Appendix: Physics Models}\label{AppendixB}

\subsection{Basic Test Models}\label{AppendixB_Basic}

Some simple models are implemented for basic tests and development, all of which are static (no time evolution):
\begin{enumerate}[label=(\roman*),itemsep=0mm]
    \item \textbf{Simple scalar fit:} The simplest model is to fit a single variable $\Phi$ to some fixed scalar field, similar to what is required for initial conditions (the 0$^{\textnormal{th}}$ timestep),
        \be
          \Phi = \Phi_0.
        \ee
    \item \textbf{Vorticity solve:} Similarly, given a scalar field $\Phi_0$, one can solve for the vorticity field
        \be
          w = \nabla^2\Phi_0.
        \ee
    \item \textbf{Vorticity with 2 variables:} Extending from this model, given a scalar field $\Phi_0$, one can solve both problems above simultaneously, such that
        \ba
          \Phi &=& \Phi_0,\\
          w &=& \nabla^2\Phi.
        \ea
        Using this model early in the development of Fuse-PINN, it became apparent that using a single model for all variables was inefficient. Effectively, the training risks getting stuck in a local minimum where $w=\Phi=0$. Separating variables forces $\Phi$ to focus on fitting $\Phi_0$, without having to compete with the loss of the vorticity equation. Once $\Phi$ becomes well resolved, the second equation has no choice but to follow.
    \item \textbf{Poisson solve:} Given a fixed scalar field $w_0$, one solves for the variable $\Phi$ with the Poisson equation
        \be
          \nabla^2\Phi = w_0.
        \ee
    \item \textbf{Poisson with 2 variables:} Extending from this model, given a fixed scalar field $w_0$, one can solve for
        \ba
          w &=& w_0,\\
          \nabla^2\Phi &=& w.
        \ea
\end{enumerate}

\subsection{Hasegawa-Wakatani}\label{AppendixB_HW}

The Hasegawa-Wakatani model solves for 2 physics variables, the density $\rho$ and the electric potential $\Phi$. However, as is the case in many tokamak-relevant fluid models, like Reduced-MHD, the Hasegawa-Wakatani model is obtained through a projection of dynamics in the plane perpendicular to the magnetic field. This projection reduces the number of velocity variables needed to describe the dynamics of the system, but it results in high-order derivatives. Therefore, it is common to use a third, auxiliary variable $w$ to reduce the order of direct derivatives needed in the system. For generality, we provide the model equations in 3D, although the toroidal components and $\partial_{\phi}$ derivatives are ignored in 2D use-cases presented in this paper.

The equations are expressed in toroidal coordinates $(R,Z,\phi)$, as follows,
\ba
    \frac{\partial\rho}{\partial t} &=& [\rho,\Phi]_p + \alpha_{\|} (\Phi - \rho) - \kappa \partial_Z\Phi + D_{\bot}\nabla_{\bot}^2\rho + D_{\|}\nabla_{\|}^2\rho \\
    \frac{\partial w}{\partial t} &=& [w,\Phi]_p + \alpha_{\|} (\Phi - \rho) + \mu_{\bot}\nabla_{\bot}^2w + \mu_{\|}\nabla_{\|}^2w \label{HW_momentum}\\
    &&\nabla_{\bot}^2\Phi \textnormal{ }=\ttt{ } w \label{HW_poisson}
\ea
where the Poisson Brackets are defined as
\be
    [a,b]_p \ttt{ }=\ttt{ } \partial_R a \ttt{ } \partial_Z b \ttt{ }-\ttt{ } \partial_R b \ttt{ } \partial_Z a \label{Poisson_brackets}
\ee
and where the perpendicular diffusion operators are defined as
\be
    \nabla_{\bot}^2 = \partial_R^2 + \frac{1}{R}\partial_R + \partial_Z^2.\label{Perp_diff_operator}
\ee
The parallel diffusion operator is aligned to the background magnetic field. In a tokamak device, the magnetic field is nearly toroidal, so its exact form can be greatly simplified in a local frame by approximating the magnetic field direction as $\mathbf{b} = \mathbf{B}/|\mathbf{B}| \approx \mathbf{e}_{\phi}$, such that
\ba
    \nabla_{\|}^2 &=& \nabla\cdot\left(\mathbf{b}[\mathbf{b}\cdot\nabla]\right) = \mathbf{b}\cdot\nabla(\mathbf{b}\cdot\nabla) \\
                 &\approx& \frac{1}{R^2}\partial_{\phi}^2 \label{Par_diff_operator}
\ea
The scalars $D_{\bot}$ and $D_{\|}$ are the perpendicular and parallel diffusion coefficients, while $\mu_{\bot}$ and $\mu_{\|}$ are the perpendicular and parallel viscosity coefficients. The coupling parameter $\alpha_{\|}$ is a simplified closure term for parallel electron dynamics, which may involve parallel derivatives, but is typically assumed to be constant. The $\kappa\partial_Z\Phi$ term stands for the drift-wave drive.

Note that the arrangement of equation-(\ref{HW_poisson}) is intentional, as we use this Poisson equation to solve for $\Phi$, whereas equation-(\ref{HW_momentum}) is used to solve for $w$.

All boundary conditions are imposed as Dirichlet with the variable values from the initial conditions. In order to initialise the system with a zero force balance in the term $\alpha_{\|} (\Phi - \rho)$, the electric potential is initialised as $\Phi=\rho$, and the vorticity is initialised accordingly with $w=\nabla_{\bot}^2\Phi$.

\subsection{Electrostatic Reduced-MHD model}\label{AppendixB_ERMHD}

The electrostatic Reduced-MHD model solves for 3 physics variables, the density $\rho$, the temperature $T$ and the electric potential $\Phi$. As in the Hasegawa-Wakatani implementation, an auxiliary variable $w$ is used to reduce the order of direct derivatives needed in the residual formulation. For generality, we provide the model equations in 3D, although the toroidal components and $\partial_{\phi}$ derivatives are ignored in 2D use-cases presented in this paper.

The equations are expressed in toroidal coordinates $(R,Z,\phi)$, as follows,
\ba
    \frac{\partial\rho}{\partial t} &=& R[\rho,\Phi]_p + 2\rho\partial_Z\Phi + D_{\bot}\nabla_{\bot}^2\rho + D_{\|}\nabla_{\|}^2\rho + S_{\rho} \\
    \frac{\partial\rho T}{\partial t} &=& R[\rho T,\Phi]_p + 2\rho T\gamma\partial_Z\Phi + K_{\bot}\nabla_{\bot}^2T + K_{\|}\nabla_{\|}^2T + S_{T} \\
    \rho R^2\frac{\partial w}{\partial t} + \nabla(\rho R^2)\cdot\nabla(\frac{\partial\Phi}{\partial t}) &=& \frac{1}{2R}[V^2,\rho R^2]_p + \frac{1}{R}[\rho R^4 w, \Phi]_p - 2\partial_Z(\rho T) + \mu_{\bot}\nabla_{\bot}^2w \label{ERMHD_momentum}\\
    &&\nabla_{\bot}^2\Phi \textnormal{ }=\textnormal{ } w \label{ERMHD_poisson}
\ea
where the Poisson Brackets $[.,.]_p$ and the perpendicular/parallel diffusion operators are defined as above with equations (\ref{Poisson_brackets}), (\ref{Perp_diff_operator}) and (\ref{Par_diff_operator}) respectively. The perpendicular velocity $V$ is defined as
\be
  V = R^2\nabla\phi\times\nabla\Phi = -(R\partial_Z\Phi)\vec{e}_R + (R\partial_R\Phi)\vec{e}_Z \label{ERMHD_velocity}
\ee
such that $V^2 = R^2 (\Phi_R^2 + \Phi_Z^2)$.

Here too, the scalars $D_{\bot}$ and $D_{\|}$ are the perpendicular and parallel diffusion coefficients, while $K_{\bot}$ and $K_{\|}$ are the perpendicular and parallel thermal conductivity coefficients, and where $\mu_{\bot}$ and $\mu_{\|}$ are the perpendicular and parallel viscosity coefficients. The ratio of specific heats is set to $\gamma=\frac{5}{3}$. The terms $S_{\rho}$ and $S_{T}$ represent density and heat sources respectively. All boundary conditions are imposed as Dirichlet with the variable values from the initial conditions.


\section{Appendix: Poisson Tests}\label{AppendixC}

The first test employs the Poisson model with a single variable $\Phi$ -- model (iv) in Appendix-\ref{AppendixB}1 -- to solve for the same field as in \cite{Muller2023}, namely $\nabla^2\Phi = -2\pi^2 sin(\pi x)sin(\pi y)$, with zero boundary conditions on a square domain with $x\in[0,1]$ and $y\in[0,1]$. This has an exact analytical solution of $\Phi = sin(\pi x)sin(\pi y)$. The test is conducted with a single PINN for the whole domain (ie. without FEM-NN) and compared against M\"uller's \& Zeinhofer's JAX implementation in \cite{Muller-github}.

The same hyperparameters are used for both codes, with the same line-search parameters, and a grid size of 100$\times$100. The size of the MLP (Multi-Layer Perceptron) Neural Network is varied to explore performance. Since there are two input dimensions $(x,y)$ and a single variable output, the input and output layers are respectively 2 and 1, and the inner layers are varied with sizes $[2,20,1]$, $[2,50,1]$, $[2,20,20,1]$, $[2,30,30,1]$ and $[2,40,40,1]$. An important ENGD parameter is the Gram Matrix regularisation, which affects how fast the solution converges, but with a trade off of how precise the solution is. This parameter is varied between $10^{-09}$ and $10^{-14}$.
\begin{figure}[ht!]
  \centering
        \subfigure[ ]{\includegraphics[height=0.28\linewidth]{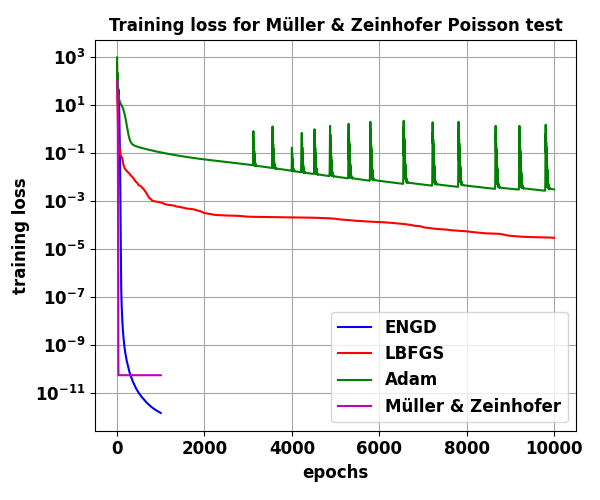}}
        \subfigure[ ]{\includegraphics[height=0.28\linewidth]{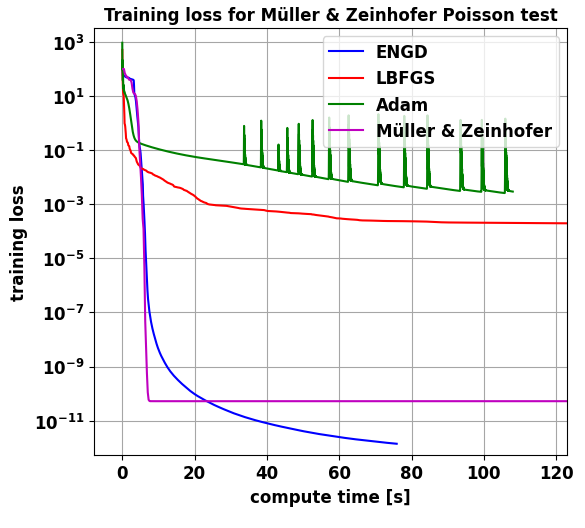}}
        \subfigure[ ]{\includegraphics[height=0.28\linewidth]{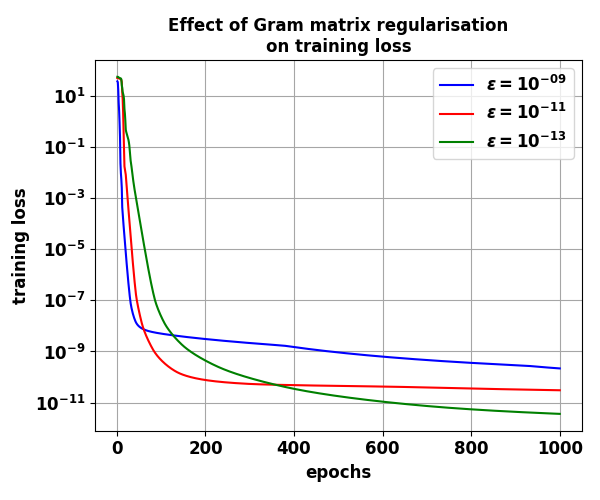}}
        \subfigure[ ]{\includegraphics[height=0.33\linewidth]{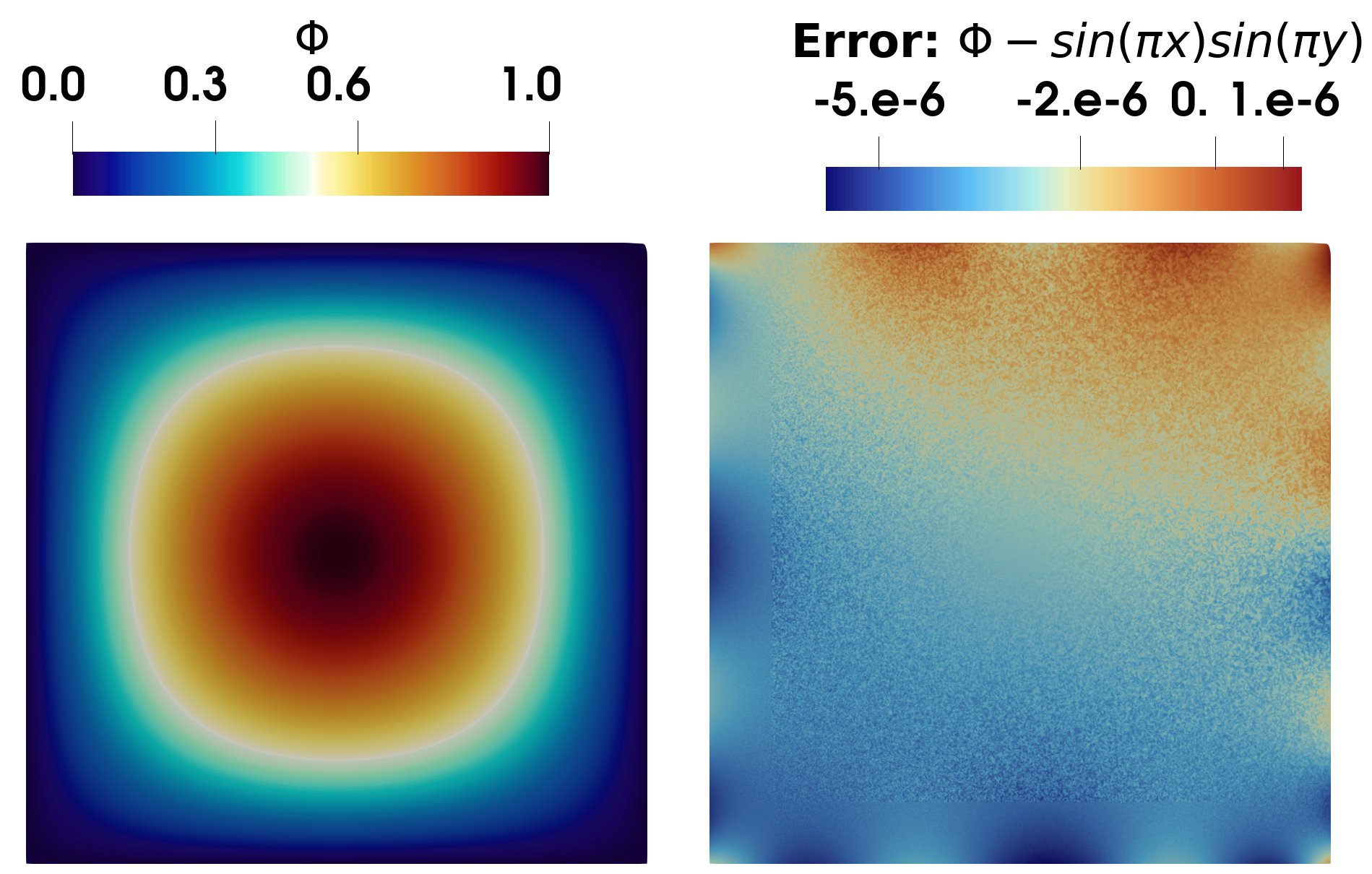}}
        \subfigure[ ]{\includegraphics[height=0.33\linewidth]{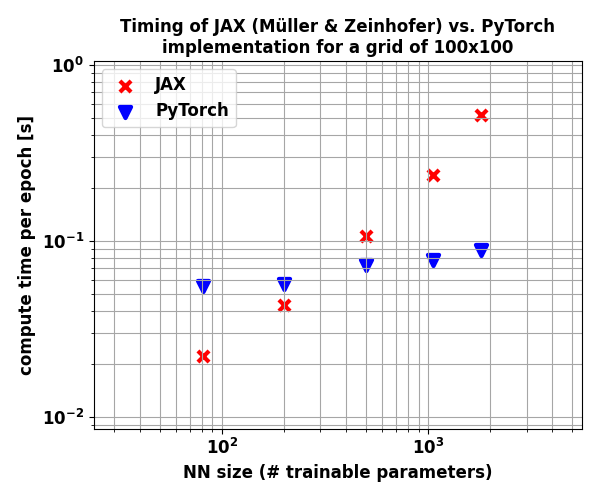}}
  \caption{Testing of Poisson equation with the same test as in the original ENGD paper \cite{Muller2023}. \textnormal{(a)} Training loss as a function of epochs, comparing ENGD with Adam and L-BFGS optimisers, as well as the original JAX implementation of ENGD from M\"uller \& Zeinhofer. \textnormal{(b)} Same training losses but as a function of compute time. \textnormal{(c)} Effect of Gram Matrix regularisation parameter: higher regularisation leads to slightly faster convergence, but with a trade-off in final convergence level. \textnormal{(d)} The 2D map of the final solution $\Phi$, together with the error compared to the analytical solution $sin(\pi x)sin(\pi y)$. \textnormal{(e)} Comparison of compute cost between the JAX implementation of the original ENGD paper \cite{Muller2023} and the PyTorch implementation of Fuse-PINN, for various NN sizes, demonstrating satisfactory performance of the Fuse-PINN implementation.}
  \label{FIGURE_Poisson}
\end{figure}

Figure-\ref{FIGURE_Poisson}a and \ref{FIGURE_Poisson}b show the training loss as a function of epochs and compute-time, respectively, and compares the ENGD performance with Adam and L-BFGS optimisers, as well as M\"uller's \& Zeinhofer's JAX version. The Fuse-PINN implementation shows similar performance as the original ENGD implementation in JAX, with slight variations at low errors thought to be due to the Gram Matrix regularisation level. As shown in Figure-\ref{FIGURE_Poisson}c, increasing the regularisation level leads to faster initial convergence, but also higher final error. A scheduled decay of the regularisation parameter could easily be implemented but was not considered essential for the purpose of this work. Figure-\ref{FIGURE_Poisson}-d shows the 2D map of the final solution, together with the error against the analytical solution.

The computation time is also similar to the original ENGD implementation in JAX, but seems to be more efficient for larger PINNs. Figure-\ref{FIGURE_Poisson}-e shows the compute time per epoch for the original ENGD implementation in JAX, compared to the Fuse-PINN implementation in PyTorch. It is not clear why the JAX implementation becomes slower with larger NN models, and there are certainly performance tuning improvements that could improve this aspect. For example, this could come from different implementations of the Least-Square routine between JAX and PyTorch. However, investigating these differences in details is beyond the scope of the present work. Potential JAX implementation of Fuse-PINN is discussed later in the limitations and future works section. Nevertheless, the comparison shown here demonstrates that the Fuse-PINN implementation is reasonably competitive and performant.

Further examples and analysis of Poisson solves are also presented in later sections, where physics models require an auxiliary variable for the vorticity $w = \nabla^2\Phi$ which is solved for $\Phi$ given much more complex fields than this test. The Poisson solve in physics models presented in further sections is typically the most challenging equation of the entire system, due to its implicit nature.

\end{document}